%% file: main.tex
\pdfoutput=1

\documentclass[11pt, dvipsnames]{article}

\usepackage[preprint]{acl}

\usepackage{times}
\usepackage{latexsym}
\usepackage{inconsolata}

\usepackage[T1]{fontenc}

\usepackage[utf8]{inputenc}

\usepackage{microtype}

\usepackage{graphicx}
\usepackage{bbm}
\usepackage{amsmath}
\usepackage{amsfonts}
\usepackage{amssymb}
\usepackage{amsthm}
\usepackage{dsfont}
\usepackage{caption}
\usepackage{subcaption}
\usepackage{soul}
\usepackage{hyperref}
\usepackage{adjustbox}
\usepackage{booktabs}
\usepackage{makecell}
\usepackage{nicefrac}
\usepackage{comment}
\usepackage{rotating}
\usepackage{scalerel}
\usepackage{lipsum}
\usepackage{svg}

\input{macros}

\title{Generalized Measures of Anticipation and Responsivity in Online Language Processing}
\author{
	Mario Giulianelli \qquad Andreas Opedal \qquad Ryan Cotterell \\
	\texttt{\{\href{mailto:mario.giulianelli@inf.ethz.ch}{mario.giulianelli},\href{mailto:andreas.opedal@inf.ethz.ch}{andreas.opedal},\href{mailto:ryan.cotterell@inf.ethz.ch}{ryan.cotterell}\}@inf.ethz.ch}\\
	\setlength{\fboxsep}{2.5pt}%
	\setlength{\fboxrule}{2.5pt}%
	\fcolorbox{white}{white}{
		\includesvg[width=.15\linewidth]{figures/ethz-logo}
	}
}

\begin{document}
	\maketitle
	\begin{abstract}
		We introduce a generalization of classic information-theoretic measures of predictive uncertainty in online language processing, based on the simulation of expected continuations of incremental linguistic contexts. Our framework provides a formal definition of anticipatory and responsive measures, and it equips experimenters with the tools to define new, more expressive measures beyond standard next-symbol entropy and surprisal. While extracting these standard quantities from language models is convenient, we demonstrate that using Monte Carlo simulation to estimate alternative responsive and anticipatory measures pays off empirically: New special cases of our generalized formula exhibit enhanced predictive power compared to surprisal for human cloze completion probability as well as ELAN, LAN, and N400 amplitudes, and greater complementarity with surprisal in predicting reading times.
		
		\vspace{.11em}
		\hspace{1.25em}\includegraphics[width=1.25em,height=1.25em]{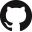}{\hspace{.75em}\parbox{\dimexpr\linewidth-2\fboxsep-2\fboxrule}{\url{https://github.com/rycolab/generalized-surprisal}}}
		
	\end{abstract}
	
	\section{Introduction}
	The prediction of upcoming linguistic units is posited to play a key role in human language comprehension \cite{federmeier2007thinking,willems2016prediction,goldstein2022shared}. 
	One fruitful method of operationalizing human uncertainty over predictions is through information-theoretic measures.
	Because human predictive mechanisms leave behavioral and neural traces that are observable during reading and listening \cite{kutas1984brain,van2005anticipating,forseth2020language}, the most common method of vetting an information-theoretic measure of predictive uncertainty is by examining its relationship with such traces. 
	Beyond simply yielding good correlates, information-theoretic measures often provide insight into the human prediction mechanism, and they are thus central to much cognitive and neurobiological research on human language processing \cite{monsalve2012lexical,armeni2017neuroscience,wilcox-2023-testing}.
	\looseness=-1 
	
	In the domain of sentence processing, there are two commonly deployed information-theoretic measures of predictive uncertainty.
	Both assume that the comprehender implicitly maintains a probability distribution over upcoming sequences of linguistic units.
	The first, and most prominent, is the surprisal of a unit given its preceding context \cite{hale2001probabilistic}, while the other is entropy \cite{Hale2003,hale2006uncertainty}.
	In broad strokes, surprisal tells us how likely the next unit is in the given context and is a good example of a \defn{responsive} measure, i.e., a measure that quantifies a response to the \emph{next} unit. 
	In contrast to surprisal, entropy is solely a function of the context, as it tells us the uncertainty over the range of possible upcoming linguistic units. 
	Thus, entropy is an example of an \defn{anticipatory} measure, i.e., a measure that anticipates a response to the next unit without knowing its identity.
	In the specific case of next-symbol entropy \cite{frank2013uncertainty,pimentel-etal-2023-anticipation}, it is the expected value of the next-symbol surprisal so it comes with a natural interpretation of the expected response. 
	
	Estimates of surprisal and entropy based on neural language models have demonstrated significant predictive capacity for a wide variety of neural and behavioral data collected using 
	self-paced and eye-tracked reading experiments \citep{goodkind2018predictive,wilcox-2023-testing}, as well as EEG \citep{merkx-frank-2021-human,michaelov2024n400}, fMRI \citep{shain2020fmri,bhattasali-resnik-2021-using}, and ECoG imaging \citep{schrimpf2021neural}, along with explicit grammaticality and acceptability ratings \cite{lau2017grammaticality,wallbridge2022investigating}.
	Despite surprisal and entropy's empirical success, there is increasing interest in defining and evaluating alternative measures.
	Examples include measures designed to disentangle different dimensions---e.g., lexical versus syntactic---of uncertainty \citep{roark-etal-2009-deriving,arehalli-etal-2022-syntactic,giulianelli-etal-2023-information} or to quantify uncertainty over spans larger than a single unit \citep{aurnhammer2019evaluating,giulianelli-etal-2024-incremental}.
	While estimating surprisal and entropy from neural language models is convenient, experimenters should chart new territory by designing their own measures, rethinking and enhancing established information-theoretic quantities to capture overlooked aspects of online language processing.
	
	In this paper, we introduce a new framework, termed generalized surprisal, for responsive and anticipatory models of online language processing.
	This framework encompasses existing information-theoretic measures as special cases and offers a new method to develop novel ones. We begin by deriving a generalization of surprisal, demonstrating that it corresponds to an expectation over continuations of a linguistic context (\cref{sec:framework}). 
	We then show how many existing measures can be seen as special cases of our generalized formula, and we propose new special cases, such as sequence-level entropy and next-symbol information value (\cref{sec:special-cases}). 
	Because some special cases cannot be calculated in closed form, we must rely on a sampling procedure. 
	This introduces a trade-off between runtime and variance, which we analyze empirically in \cref{sec:analysis}.
	Finally, we evaluate all special cases as predictors of neural and behavioral data collected in experiments with human participants (\cref{sec:predictive-power}).
	
	We present several new findings, including:
	(1) contextual probability predicts human cloze completions better than surprisal, while surprisal is a better predictor of human predictability ratings;
	(2) information value predicts N400 better than surprisal, which is among the go-to predictors for this ERP component \cite{delong-2005-probabilistic,frank2015erp,michaelov2024n400};
	(3) sequence-level entropy, introduced in this paper, is the sole significant predictor of ELAN; and
	(4) different responsive measures predict ERP amplitudes at varying time windows after stimulus onset.\looseness-1

	\section{Generalized Surprisal}
	\label{sec:framework}
	This section introduces our framework.
	First, we establish some key notation and definitions.
	Then, we present a decomposition of surprisal, which motivates our definition of generalized surprisal.
	
	\subsection{Language Modeling}
	\label{sec:framework-definitions}
	An \defn{alphabet} $\alphabet$ is a finite, non-empty set of symbols, and its Kleene closure $\kleene{\alphabet}$ is the set of all strings formed by concatenating symbols in $\alphabet$, including the empty string $\varepsilon$.\footnote{
		We use an unbolded font for symbols, i.e., $\sym \in \alphabet$, and a bolded font for strings $\str \in \kleene{\alphabet}$.
		The concatenation of two strings $\str$ and $\continuation$ is written as $\str\continuation$. 
		The length of a string is the number of symbols it contains and is denoted as $|\str|$.}
	The set of all strings $\kleene{\alphabet}$ is partially ordered by the \defn{prefix relation} $\preceq$, defined as follows: $\str \preceq \strpr \iff \exists \continuation: \str\continuation = \strpr$.
	As is easy to see, $\preceq$ is reflexive and transitive, but not symmetric.
	A \defn{language model} $\pLM$ is a distribution over strings $\kleene{\alphabet}$. 
	A common quantity derived from a language model is the \defn{prefix probability}, defined as 
	\begin{equation}\label{eq:prefix-prob}
		\paLM(\str \mid \ctx) \defeq \sum_{\continuation \in \kleene{\alphabet}} \pLM(\str \continuation  \mid \ctx).
	\end{equation}
	In words, \Cref{eq:prefix-prob} tells us the probability of the event that a string sampled from $\pLM$ starts with $\str$. 
	Crucially, this is different from the probability $\pLM(\str \mid \ctx)$ that the string \emph{is} identically $\str$.\looseness=-1

	\paragraph{The Human Language Model.}
	So far, we have used the symbol $\pLM$ to refer to an arbitrary language model.
	However, in the context of cognitive modeling, we are interested in a hypothetical construct model---the human language model $\pHum$.
	Because the true human language model is unknown, we must approximate it via another language model $\qLM$.
	To the extent that $\qLM$ is close to $\pHum$ (under some notion of distance between distributions), we would expect estimates derived from $\qLM$ to be a reliable proxy of the probabilities prescribed by the human language model.
	In our experiments, we will use a model $\qLM$ parameterized by a transformer neural network, which was shown to closely approximate $\pHum$ in a series of psycholinguistic studies \cite[][\textit{inter alia}]{schrimpf2021neural,oh2022does,shainetal24}.\looseness=-1
	
	\subsection{Generalizing Surprisal}
	\label{sec:framework-generalizing}
	The \defn{surprisal} of a \defn{target} $\str \in \kleene{\alphabet}$ given a \defn{context} $\ctx \in \kleene{\alphabet}$ is defined as $\surprisal_{\qLM}(\str ; \ctx) \defeq -\log \qaLM(\str \mid \ctx)$.\footnote{
		Note that this is not equal to information content in a strict sense, since that would require $\qaLM(\cdot \mid \ctx)$ to be a probability distribution over $\kleene{\alphabet}$, for all $\ctx \in \kleene{\alphabet}$. 
	}
	In constructing our framework, we draw inspiration from the following decomposition of surprisal:
	\begin{subequations}
		\begin{align} 
			\surprisal_{\qLM}(\str ; \ctx) &\defeq-\log \qaLM(\str \mid \ctx) \label{eq:surprisal-decomposed} \\
			&=- \log \sum_{\continuation \in  \kleene{\alphabet}}  \qLM(\str \continuation \mid \ctx)  \\
			&=- \log \sum_{\continuation \in  \kleene{\alphabet}}  \qLM(\continuation \mid \ctx)  \indicator{\str \preceq \continuation},
		\end{align}
	\end{subequations}
	where $\continuation$ is a \defn{continuation} and $\indicator{\str \preceq \continuation}$ is an indicator function that returns $1$ when $\str \preceq \continuation$ is true and $0$ otherwise.
	When viewed through the lens of such a decomposition, surprisal is a marginalization over possible continuations of the context $\ctx$: it is the negative log-transformed cumulative probability of all the continuations that begin with $\str$.\looseness=-1
	
	By writing the sum as an expectation, we can rewrite surprisal as follows
	\begin{equation}
		\surprisal_{\qLM}(\str ; \ctx) = - \log \left( \Expect_{\continuation \sim \qLM(\cdot \mid \ctx)} \left[ \indicator{\str \preceq \continuation} \right]\right).
		\label{eq:generalized-surprisal}
	\end{equation}
	Our notion of generalized surprisal abstracts \Cref{eq:generalized-surprisal} by
	introducing a \defn{scoring function} $\funcg$ to generalize the indicator function and a \defn{warping function} $\funcf$ to replace the $-\log$ of standard surprisal.
	\begin{definition}[Generalized Surprisal]\label{def:generalized-surprisal}
		We define a \defn{generalized surprisal} model as the pair $(\funcf, \funcg)$ of a warping function $\funcf \colon \mathbb{R} \to \mathbb{R}$ and a scoring function $\funcg \colon \kleene{\alphabet} \times \kleene{\alphabet} \times \kleene{\alphabet} \to \mathbb{R}$.
		Under a specific model $(\funcf, \funcg)$, the generalized surprisal of a target $\str$ in a context $\ctx$ is\looseness=-1
		\begin{equation}
			\wordmeasure_{\qLM}(\str; \ctx) \defeq \funcf \left( \Expect_{\continuation \sim \qLM(\cdot \mid \ctx)} \left[ \funcg(\continuation, \str, \ctx)\right]\right).
			\label{eq:gen-surprisal}
		\end{equation}
	\end{definition}
	
	\paragraph{The Scoring Function.}
	We call $\funcg$ the \emph{scoring} function because it evaluates each continuation $\continuation \sim \qLM(\cdot \mid \ctx)$ against a target~$\str$, conditioned on a context~$\ctx$, yielding a real-valued score.
	The score quantifies the accuracy of a prediction (or how close the prediction is to the observed target), where the specific notion of closeness is encoded by the experimenter in their definition of~$\funcg$.
	
	\paragraph{The Warping Function.}
	We call $\funcf$ the \emph{warping} function because it applies a transformation to the expected score and thus controls the shape of the distribution of generalized surprisal values for a given score distribution. 
	It is useful to think of the warping function as characterizing the relationship between prediction accuracy and a certain construct or measurement of interest.
	For instance, in \cref{sec:results-responsive}, we show how the same notion\ryan{Why is it binary? I think it's still in $[0, 1]$ rather than $\{0, 1\}$. \response{Mario} It is in $[0, 1]$ in expectation, but the scoring function itself is binary, do you agree?} of prediction accuracy captured by surprisal's scoring function, $\indicator{\str \preceq \continuation}$, is in a nearly linear relationship with human cloze probabilities yet in a logarithmic relationship with human predictability ratings.
	Much psycholinguistic research aiming to establish the functional relationship between surprisal and processing difficulty \cite[][\textit{inter alia}]{smith2013-log-reading-time,brothers-kuperberg-2021,wilcox-2023-testing,shainetal24} can be seen as testing different hypotheses about the workings of online language processing by instantiating them through varying warping functions.\looseness-1

	\subsection{Anticipation and Responsivity}
	\label{sec:framework-formula}
	An important distinction between various generalized surprisal models is whether they characterize anticipatory or responsive online processes.
	These notions have been introduced informally by \citet{pimentel-etal-2023-anticipation}. 
	We give a formal definition of anticipation and responsivity below.
	
	\begin{definition}[Anticipation and Responsivity]\label{def:anticipation-and-responsivity}
		We call a generalized surprisal model $(\funcf, \funcg)$ \defn{anticipatory} if $\funcg$ is constant in $\str$, i.e., $\forall \continuation,\str,\str',\ctx \in \kleene{\alphabet}$, we have $\funcg(\continuation,\str, \ctx) = \funcg(\continuation, \str', \ctx)$. 
		Otherwise, we call $(\funcf, \funcg)$ \defn{responsive}.
	\end{definition}
	\cref{def:anticipation-and-responsivity} 
	differentiates anticipation, a state of uncertainty over possible outcomes that is fully determined by the context and the processor's language model, from responsivity, which expresses uncertainty for a specific next outcome.
	
	\section{Special Cases of Generalized Surprisal}\label{sec:special-cases}\label{sec:generalized-surprisal-models}
	In this section, we introduce concrete special cases of generalized surprisal (\cref{eq:gen-surprisal}), which we evaluate as predictors of human behavior and neural activity recorded during online language processing. 
	Some of these have been previously used to predict such psycholinguistic data, while others are new. 
	All special cases are designed by varying the three core components of our framework---anticipation vs.\ responsivity, scoring function, and warping function---and are meant to exemplify how different hypotheses about online language processing can be instantiated as generalized surprisal models.\looseness-1
	
	\subsection{Responsive Measures}\label{sec:responsive}
	We start by introducing three responsive generalized surprisal models. 
	These are models $(\funcf, \funcg)$ where the scoring function $\funcg(\continuation,\str, \ctx)$ is \textit{not} constant in $\str$ (see \cref{def:anticipation-and-responsivity}). 
	
	\paragraph{Surprisal.}
	The first generalized surprisal model we will consider is the pair $(\funcf, \funcg)$ corresponding to standard surprisal (\cref{eq:surprisal-decomposed,eq:generalized-surprisal}):
	\vspace{-0.8em}
	\boxalign{
		\begin{subequations}
			\begin{align}
				\funcf(x) &= - \log (x) \label{eq:surprisal-model-f}\\
				\funcg(\continuation, \str, \ctx) 
				&= \indicator{\str \preceq \continuation}. \label{eq:surprisal-model-g}
			\end{align}
		\end{subequations}
	}
	The scoring function captures a binary notion of prediction accuracy, while the logarithmic warping function is classically considered to instantiate a view of online language processing where cognitive costs reflect the magnitude of incremental mental representation updates, which is related logarithmically to prediction accuracy \cite{levy2008expectation}.
	
	\paragraph{Probability.}
	Replacing standard surprisal's logarithmic warping function with the identity function yields the contextual probability of the next unit:\looseness=-1
	\vspace{-0.8em}
	\boxalign{
		\begin{subequations}
			\begin{align}
				\funcf(x) &= x \label{eq:probability-model-f}\\
				\funcg(\continuation, \str, \ctx) &= \indicator{\str \preceq \continuation}. \label{eq:probability-model-g}
			\end{align}
		\end{subequations}
	}
	The identity warping function can be seen as instantiating a \textit{facilitation} view of online linguistic prediction, where prediction accuracy is linearly related to cognitive cost \citep{brothers-kuperberg-2021}.\looseness-1
	
	\paragraph{Information Value.}
	By replacing the indicator function of surprisal and probability with a scoring function $\distance_{\ctx} \colon \kleene{\alphabet} \times \kleene{\alphabet} \to \mathbb{R}_{\geq 0}$ that measures a graded, possibly context-sensitive notion of distance between strings, we obtain information value \cite{giulianelli-etal-2023-information}:\footnote{
		More precisely, this is information value with the \textit{mean} as a summary statistic \cite[cf.][\S3.1]{giulianelli-etal-2023-information}. The ordered pair $(\kleene{\alphabet}, \distance_{\ctx})$ forms a semi-metric space, satisfying all properties of a metric space except, possibly, the triangle inequality. The distance function $\distance_{\ctx}$ may also be constant in $\ctx$.\looseness-1
	}
	\vspace{-0.8em}
	\boxalign{
		\begin{subequations}
			\begin{align} 
				\funcf(x) &= x \\
				\funcg(\continuation, \str, \ctx) &= \distance_{\ctx}(\continuation, \str).
			\end{align}
		\end{subequations}
	}
	Whilst the use of a binary scoring function follows naturally from the surprisal model of cognitive cost, it results in a relatively simplistic notion of prediction accuracy which conflates different aspects of predictive accuracy and does not take into account the communicative equivalence of predictions and observations.
	In the information value model of cost \cite{giulianelli-etal-2023-information,giulianelli-etal-2024-incremental}, instead, prediction accuracy is a continuous score that quantifies the representational distance between predictions and observations.
	For instance, if the predicted continuation is syntactically different but semantically equivalent to the observed next unit, this results in high syntactic and low semantic information value.

	\subsection{Anticipatory Measures} \label{sec:anticipatory}
	We now move to anticipatory generalized surprisal models, where the scoring function $\funcg(\continuation,\str, \ctx)$ is constant in $\str$; see \cref{def:anticipation-and-responsivity}.
	We are not aware of any theoretical justifications for using non-linear warping functions for anticipatory measures, so all the special cases presented here will use the identity function $\funcf(x) = x$.
	
	\paragraph{Expected Next-symbol Surprisal.}
	We begin with a measure that was recently proposed to study the effects of anticipatory processing on reading comprehension \citep{pimentel-etal-2023-anticipation}. 
	This is the expected surprisal over the language model's next-symbol distribution, which is defined as follows:\looseness-1\footnote{
		Note that this expression involves $\qaLM(\cdot \mid \ctx)$ rather than $\qLM(\cdot \mid \ctx)$ since it quantifies the uncertainty over the \emph{first} symbols of possible continuations, rather than over full continuations. The $\qLM(\varepsilon \mid \ctx)$ term is included to account for the possibility that the string ends after $\ctx$, i.e., that the continuation \emph{is} $\varepsilon$.\looseness-1
	}
	\begin{equation}
		\begin{aligned}
			- \sum_{\symu \in \alphabet} &\qaLM(\symu \mid \ctx) \log \qaLM(\symu \mid \ctx) \\
			&\qquad \quad - \qLM(\varepsilon \mid \ctx) \log \qLM(\varepsilon \mid \ctx).
		\end{aligned}
	\end{equation}
	This measure can be obtained by instantiating the following generalized surprisal model:
	\vspace{-0.8em}
	\boxalign{
		\begin{subequations}
			\begin{align}
				\funcf(x) &= x \\
				\funcg(\continuation, \str, \ctx) 
				&= - \sum_{\symu \in \alphabet} \indicator{\symu \preceq \continuation} \log  \qaLM(\symu \mid \ctx) \nonumber  \\
				& \,\,\quad\quad -  \indicator{\varepsilon = \continuation} \log \qLM(\varepsilon \mid \ctx).
			\end{align}
		\end{subequations}
	}
	This model lends itself to multiple interpretations; see \citet[\S3]{pimentel-etal-2023-anticipation} for some proposals.
	One prominent view is that cognitive resources may be budgeted in advance---before the identity of the next symbol is known---in proportion to the magnitude of the mental representation update the processor expects to sustain once the symbol is observed. As we have seen in \cref{sec:responsive}, said magnitude corresponds to the logarithm of the prediction accuracy.\looseness-1

	\paragraph{Expected Next-symbol Probability.}
	An alternative view of anticipatory mechanisms is that they allow for preemptive processing of upcoming units, with cognitive costs proportional to the prediction accuracy \textit{in string space}.
	As seen in \cref{sec:responsive}, this view can be expressed by discarding the logarithm:\looseness-1
	\vspace{-0.8em}
	\boxalign{
		\begin{subequations}
			\begin{align}
				\funcf(x) &= x \\
				\funcg(\continuation, \str, \ctx) &=\sum_{\symu \in \alphabet} \indicator{\symu \preceq \continuation} \qaLM(\symu \mid \ctx) \nonumber  \\
				& \,\,\quad\quad  +\indicator{\varepsilon = \continuation} \qLM(\varepsilon \mid \ctx).
			\end{align}
		\end{subequations}
	}
	Under this model, expected prediction accuracy is linearly related to cognitive cost.
	
	\paragraph{Expected Next-symbol Information Value.}
	A third view is that contextual uncertainty increases processing cost by requiring the retention of a larger number of competing continuations in memory.
	While this can be modeled through expected next-symbol surprisal and probability \cite{pimentel-etal-2023-anticipation}, these measures characterize different continuations as
	distinct objects.
	An alternative hypothesis is that maintaining multiple continuations in memory should be less costly if they are representationally similar.
	This can be expressed through information value's scoring function:
	\vspace{-0.8em}
	\boxalign{
		\begin{subequations}
			\begin{align}
				\funcf(x) &= x \\
				\funcg(\continuation,\str, \ctx) &= \Expect_{\continuation' \sim \qLM(\cdot \mid \ctx)} \distance_{\ctx}(\contpos{1}, \contposprime{1}),
			\end{align}
		\end{subequations}
	}
	with a representational notion of distance $\distance_{\ctx}$.\looseness=-1
	
	\paragraph{Entropy.}
	The three anticipatory models presented so far assume that anticipatory processes operate solely over the next symbol.
	One way to capture uncertainty over sequences of symbols is through the entropy of the language model, i.e.,
	\begin{equation}
		-\sum_{\continuation \in \kleene{\alphabet}} \qLM(\continuation \mid \ctx) \log \qLM(\continuation \mid \ctx).
	\end{equation}
	The corresponding generalized surprisal model is
	\vspace{-0.8em}
	\boxalign{
		\begin{subequations}
			\begin{align}
				\funcf(x) &= x \\
				\funcg(\continuation, \str, \ctx) &= - \log \qLM(\continuation \mid \ctx).
			\end{align}
		\end{subequations}
	}
	This model inherits the interpretation of expected surprisal but characterizes anticipation as a process spanning longer time intervals \cite{Hale2003}.
	
	\paragraph{Expected Information Value.}
	We also consider a second generalized surprisal model that captures sequence-level contextual uncertainty:
	\vspace{-0.8em}
	\boxalign{
		\begin{subequations}
			\begin{align}
				\funcf(x) &= x \\
				\funcg(\continuation, \str, \ctx) &= \Expect_{\continuation' \sim \qLM(\cdot \mid \ctx)}  \distance_{\ctx}(\continuation, \continuation'),
			\end{align}
		\end{subequations}
	}
	again, with a representational notion of distance $\distance_{\ctx}$.
	Unlike entropy, which regards expected next strings as distinct,
	expected information value \cite[\S G]{giulianelli-etal-2023-information} corrects for potential similarity between strings in representational space.
	
	\subsection{Monte Carlo Simulation}
	
	\label{sec:monte-carlo}
	For tractable estimation of generalized surprisal, which requires taking an expectation over continuations $\continuation \in \kleene{\alphabet}$, we use Monte Carlo simulation:
	\begin{gather}
		\estwordmeasure_{\qLM}(\str; \ctx) = \funcf \left( \frac{1}{\samplesize} \sum_{\sampleidx=1}^\samplesize \funcg(\continuation^{(\sampleidx)}, \str, \ctx) \right),\label{eq:monte-carlo-estimator}
	\end{gather}
	where $\continuation^{(\sampleidx)} \sim \qLM(\cdot \mid \ctx)$ are obtained via ancestral sampling.
	If $\funcf$ is continuous, then \Cref{eq:monte-carlo-estimator} is consistent.
	Moreover, if $\funcf(x) = x$, the estimator is additionally unbiased. 
	The variance of \Cref{eq:monte-carlo-estimator} depends on the scoring function $\funcg$ and the sample size $\samplesize$; in \cref{sec:analysis-variance}, we study the influence of $\samplesize$ on the variance of the estimator for different generalized surprisal models.
	
	Some special cases of generalized surprisal---in particular, surprisal, probability, and their expected next-symbol versions---can be computed in closed form, and, for those, we do not need to rely on Monte Carlo simulation.\looseness=-1
	
	\section{Experimental Setup}

	\paragraph{Dataset.}
	\label{sec:exp-setup-data}
	We use the Aligned dataset \cite{devarda-etal-2023}, which consists of $\datasetsize = 1726$ target--context pairs from English novels annotated with several different neural and behavioral measurements. We include most of these in our experiments: cloze completions (probability and entropy), predictability ratings, event-related brain potentials (ELAN, LAN, N400, P600, EPNP, PNP), eye-tracked reading times (first-fixation time, first-pass time, right-bounded time), and self-paced reading times. Details on these measurements are given in \cref{sec:aligned-data}.
	Each target--context pair $(\str^{(\datasetidx)}, \ctx^{(\datasetidx)})$, termed a \defn{stimulus}, is associated with a real-valued measurement $\datum(\str^{(\datasetidx)}, \ctx^{(\datasetidx)})$, termed \defn{datum}, which is an aggregation of per-subject measurements for that stimulus.\footnote{
		For all types of psycholinguistic data except cloze probability and cloze entropy (which are aggregates by definition), a single datum is the average per-subject measurement. 
	}
	In our experiments, we will compute generalized surprisal $\{\estwordmeasure
	_{\qLM}(\str^{(\datasetidx)}, \ctx^{(\datasetidx)})\}_{\datasetidx=1}^{\datasetsize}$ for all stimuli in the dataset, and evaluate it as a predictor for the corresponding data $\{\datum(\str^{(\datasetidx)}, \ctx^{(\datasetidx)})\}_{\datasetidx=1}^{\datasetsize}$.
	The contexts $\ctx$ are strings ranging from 5 to 14 words and targets $\str$ are strings corresponding to a single word.\footnote{ 
		A word is taken to be a contiguous sequence of characters delimited by a white space.
		Following \citet{devarda-etal-2023}, sentence-initial and sentence-final words, as well as words attached to a comma or clitics are excluded.
	}\looseness=-1 
	
	\begin{figure*}[h!]
		\centering
		\includegraphics[width=0.95\linewidth]{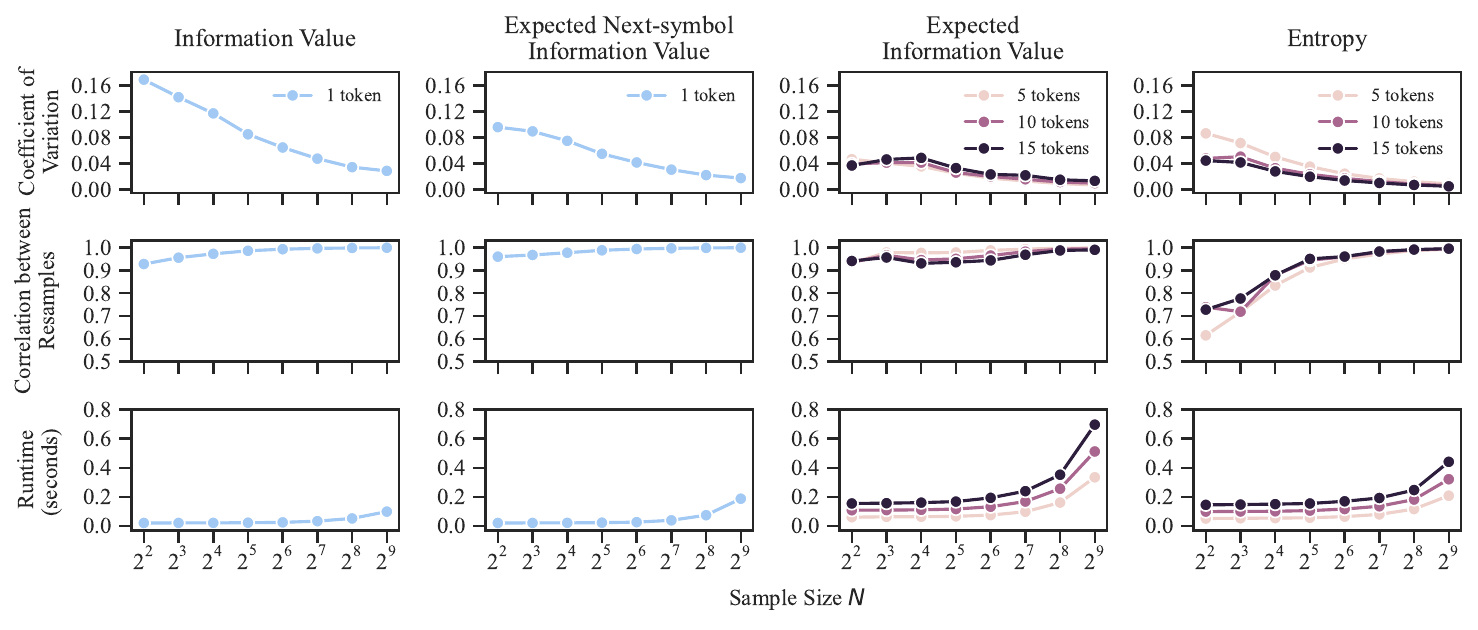}
		\vspace{-0.7em}
		\caption{\textit{Coefficient of variation} (top), \textit{correlation between resamples} (center), and \textit{runtimes} (bottom) for sampling-based measures across the stimuli in the Aligned dataset.
			Confidence intervals ($95\%$) are too narrow to be visible; the horizontal axis is in log scale.
			The average runtime for the exact metrics (surprisal, probability, expected next-symbol surprisal, and expected next-symbol probability) is $0.002$ seconds.}
		\label{fig:analysis-all-in-one-gpt2}
		\vspace{-0.8em}
	\end{figure*}
	
	\paragraph{Language Models.}
	We obtain generalized surprisal estimates from   
	\gpttwo{} \cite{radford2019language} and \gptneo{} \cite{gpt-neo}.
	Prior work has shown that, despite exhibiting higher test perplexity, these two models have better psycholinguistic predictive power than larger ones \cite{oh2022does,shainetal24}. 
	Furthermore, their smaller size incurs a lower computational cost for sampling.
	For comparability to prior work  \cite[][\textit{inter alia}]{wilcox-2023-testing,pimentel-etal-2023-anticipation,devarda-etal-2023}, when a word is composed of multiple subword tokens, token-level estimates of generalized surprisal are aggregated (either summed or multiplied, depending on the special case; see \Cref{sec:appendix-multi-token} for details).

	\paragraph{Distance Function.}
	The information value models introduced in \cref{sec:responsive} rely on a function $\distance_{\ctx}$ to quantify the distance between two strings $\continuation$ and $\str$ in the context of $\ctx$.
	Here, we use mean-pooled non-contextual (i.e., layer 0) representation from \gpttwo{} and \gptneo{} as column vectors $\repr(\continuation),\repr(\str) \in \mathbb{R}^{\embeddim}$, where $\embeddim$ is the dimensionality of the representations, 
	and calculate the cosine distance $\distance_{\ctx}(\continuation, \str) = 1 -\frac{\repr(\continuation)^{\intercal} \repr(\str)}{\|\repr(\continuation)\|_2 \|\repr(\str)\|_2}$.
	There exist several other choices for distance and representation functions; see \citet{giulianelli-etal-2023-information,giulianelli-etal-2024-incremental} and \citet{meister-etal-2024-similarity} for some examples.\looseness-1

	\section{Empirical Analysis of Measures}
	\label{sec:analysis}
	We begin our experiments with an empirical analysis of Monte Carlo estimation for the measures presented in \cref{sec:special-cases}.
	As alluded to earlier, sampling-based measures introduce variance that does not arise for measures like surprisal, which can be computed in closed form. 
	This variance can be reduced by increasing the sample size (\cref{sec:analysis-variance}), but that incurs additional costs in terms of runtime (\cref{sec:analysis-runtime}). 
	In this section, we provide an empirical analysis of these properties to gain an understanding of their trade-off. 
	We also investigate the correlations between different measures, in \cref{sec:analysis-correlations}, to assess the extent to which various theoretical models of anticipatory and responsive processing lead to divergent empirical uncertainty measurements.
	The analysis in this section is based on \gpttwo{}; results for \gptneo{} are provided in Appendix \cref{sec:appendix-variance-runtime}.

	\subsection{Variation in Estimates}
	\label{sec:analysis-variance}
	We use bootstrapping \citep{efron1992bootstrap} to measure the variance of different estimators.
	For each stimulus $(\str^{(\datasetidx)}, \ctx^{(\datasetidx)})$ in the Aligned dataset (\cref{sec:exp-setup-data}), given an original sample of size $\samplesize \in \{2^j \mid j=2,3,\dots,9\}$, we obtain $\bootstrapsamples = 1000$ resamples of the same size by sampling \emph{with replacement}.
	For entropy and expected information value, to yield tractable estimation, we limit the maximum length in tokens $\maxseqlen \in \{5,10,15\}$ of sampled continuations.
	
	In a first analysis, we compute $\mu_{\datasetidx}$ and $\sigma_{\datasetidx}$ as the mean and standard deviation of a measure of choice across the $\bootstrapsamples$ resamples and calculate the coefficient of variation $CV_{\datasetidx} = \nicefrac{\sigma_{\datasetidx}}{\mu_{\datasetidx}}$.
	\Cref{fig:analysis-all-in-one-gpt2} (top) shows how average $CV$, as expected, decreases with $\samplesize$.
	The maximum sample length $\maxseqlen$ has a limited effect on the $CV$ of sequence-level measures. 
	In a second analysis, to gauge the robustness of a given measure across a dataset of stimuli, we also calculate correlations between different resamples.
	For each measure, we obtain a matrix of $ \datasetsize \times \bootstrapsamples$ estimates, and then compute a vector of ${\bootstrapsamples \choose 2}$ Pearson correlations between all two-column combinations.
	\cref{fig:analysis-all-in-one-gpt2} (center) shows the average correlation coefficients as a function of increasing sample sizes.
	With the exception of entropy, which only reaches near-perfect correlation with $\samplesize=2^7$, all measures show near-perfect correlation already with $\samplesize=2^5$, indicating that all bootstrapped resamples yield virtually equivalent estimates for the dataset.\looseness-1
	
	In sum, for all measures, both stimulus- and corpus-level variance are contained, they decrease with larger sample sizes, and a sample size as small as $2^5$ yields consistent estimates.\looseness-1
	
	\subsection{Runtime}
	\label{sec:analysis-runtime}
	We study runtime using the same sample sizes $\samplesize$ and maximum sequence lengths $\maxseqlen$ as above, and for the same measures. \cref{fig:analysis-all-in-one-gpt2} (bottom) displays the results. 
	As expected, the runtime increases monotonically in both $\samplesize$ and $\maxseqlen$. Comparing these results to those of the variance analysis suggests a good runtime-variance tradeoff can be obtained with a sample size $\samplesize$ between $2^5$ and $2^7$.
	
	\section{Psycholinguistic Predictive Power}
	\label{sec:predictive-power}
	We now evaluate the generalized surprisal models introduced in \cref{sec:special-cases} in terms of their predictive power for the neural and behavioral data presented in \cref{sec:exp-setup-data}.
	
	\begin{figure*}[t!]
		\centering
		\includegraphics[width=0.99\linewidth]{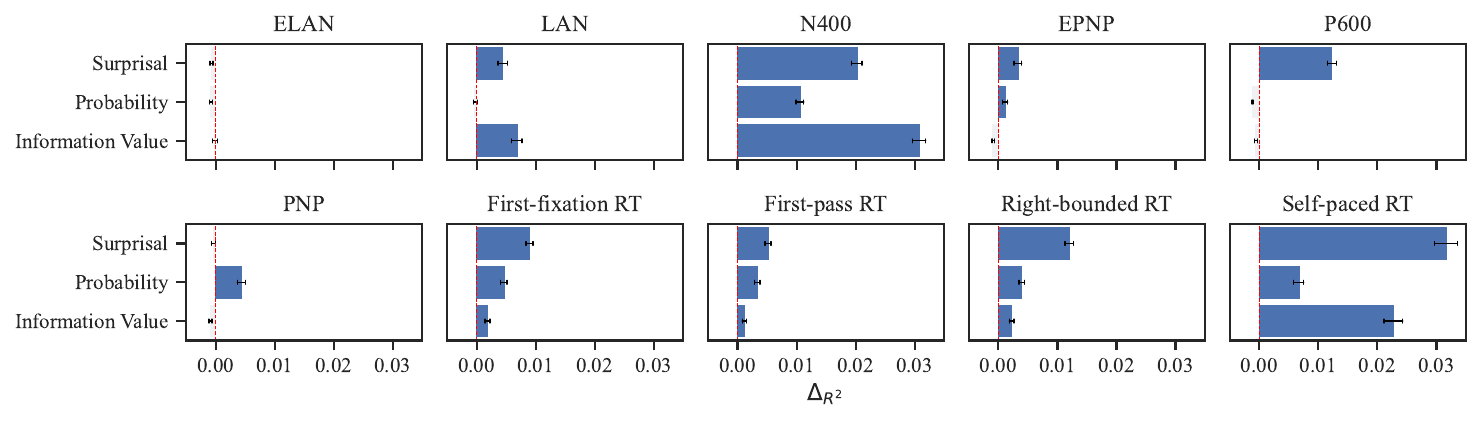}
		\vspace{-1.0em}
		\caption{Predictive power ($\deltarsq$) of responsive generalized surprisal models for event-related potentials and reading times. 95\% confidence intervals. Significance color-coded: blue for $p<0.0001$, gray for $p>0.01$.\looseness-1}
		\vspace{-0.8em}
		\label{fig:pred-power-responsive}
	\end{figure*}
	
	\begin{figure}[t]
		\centering
		\includegraphics[width=0.9\linewidth]{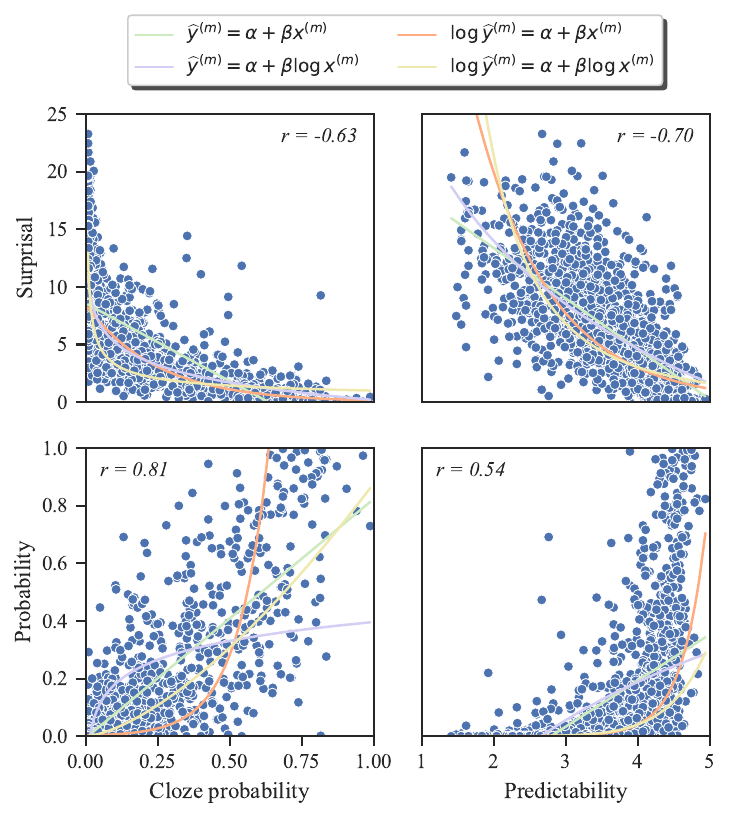}
		\vspace{-0.5em}
		\caption{Probability and surprisal against human cloze probabilities and predictability ratings, with Pearson correlation coefficients $r$ and regression lines. For regressions, $x^{(\datasetidx)} =  \wordmeasure(\str^{(\datasetidx)}, \ctx^{(\datasetidx)})$ is the predictor and $y^{(\datasetidx)} = \datum(\str^{(\datasetidx)}, \ctx^{(\datasetidx)})$ the predicted variable. 
		}
		\label{fig:surprisal_prob_vs_ratings}
		\vspace{-0.8em}
	\end{figure}
	
	\subsection{Evaluation}
	\label{sec:predictive-power-procedure}
	To quantify a measure's predictive power for a given data type $\{\datum(\str^{(\datasetidx)}, \ctx^{(\datasetidx)})\}_{\datasetidx=1}^{\datasetsize}$ in the Aligned dataset, we use regression analysis.
	We compare a regressor that includes baseline predictors for $\{(\str^{(\datasetidx)}, \ctx^{(\datasetidx)})\}_{\datasetidx=1}^{\datasetsize}$, the \defn{baseline regressor}, to one that further includes the measure of interest $\{\estwordmeasure_{\qLM}(\str^{(\datasetidx)}, \ctx^{(\datasetidx)})\}_{\datasetidx=1}^{\datasetsize}$, the \defn{target regressor}.
	Reading time regressors include target and baseline predictors not just for the target string but also the previous two words to account for spillover effects \cite{just1982paradigms,frank2013reading}.\looseness-1
	
	For the experiments on responsive measures (\cref{sec:results-responsive}), all regressors include three baseline predictors: the length of the target string $\str^{(\datasetidx)}$, its frequency, and the length of the context string $\ctx^{(\datasetidx)}$.\footnote{The length of the target string is measured in characters, the length of the context in words. 
		Frequencies are extracted from the SUBTLEXus \citep{brysbaert2012adding}.}
	We call this the \defn{default baseline}.
	For the experiments on anticipatory measures (\cref{sec:results-anticipatory}), we use an additional baseline. 
	Because expected next-symbol surprisal has proven to be most effective as a predictor when used in conjunction with surprisal \citep{pimentel-etal-2023-anticipation}, we compare the other, as yet untested, anticipatory measures against a baseline that includes both surprisal and expected next-symbol surprisal, along with the three baseline predictors mentioned above (the \defn{combined baseline}). 
	In the target regressor, expected next-symbol surprisal is then replaced with another anticipatory target predictor, allowing us to assess the boost, or decrease, in predictive power that results from this substitution.\looseness-1\footnote{
		\Cref{sec:appendix-predictive-power-results} also reports results for the default baseline.
	}
	
	In our analysis, we only use linear regression so that the functional\ryan{I would have a nice footnote relating the warpoing function to, say, GAMs here.\response{Mario}{How about this?}} relationship between the scoring function and the psycholinguistic variable can be unambiguously expressed via the warping function.\footnote{
		Others \citep[e.g.,][]{smith2008optimal,smith2013-log-reading-time,goodkind2018predictive,brothers-kuperberg-2021,wilcox-2023-testing} have tried to learn the form of this relationship from data, using generalized additive models \cite{wood-2004-stable,wood2017generalized}.
	}
	The regressors are fit with the ordinary least squares method and evaluated on a held-out test set.
	We quantify the \defn{predictive power} of a measure as the difference in the coefficient of determination $\rsq$ of the target regressor and the baseline regressor, denoted as $\deltarsq$.
	
	The statistical significance of a measure's $\deltarsq$ is assessed via 10-fold cross-validation and permutation tests.
	A full description of our analysis procedure is given in \cref{sec:appendix-significance-testing}.\looseness-1

	\begin{figure*}
		\includegraphics[width=1.01\linewidth]{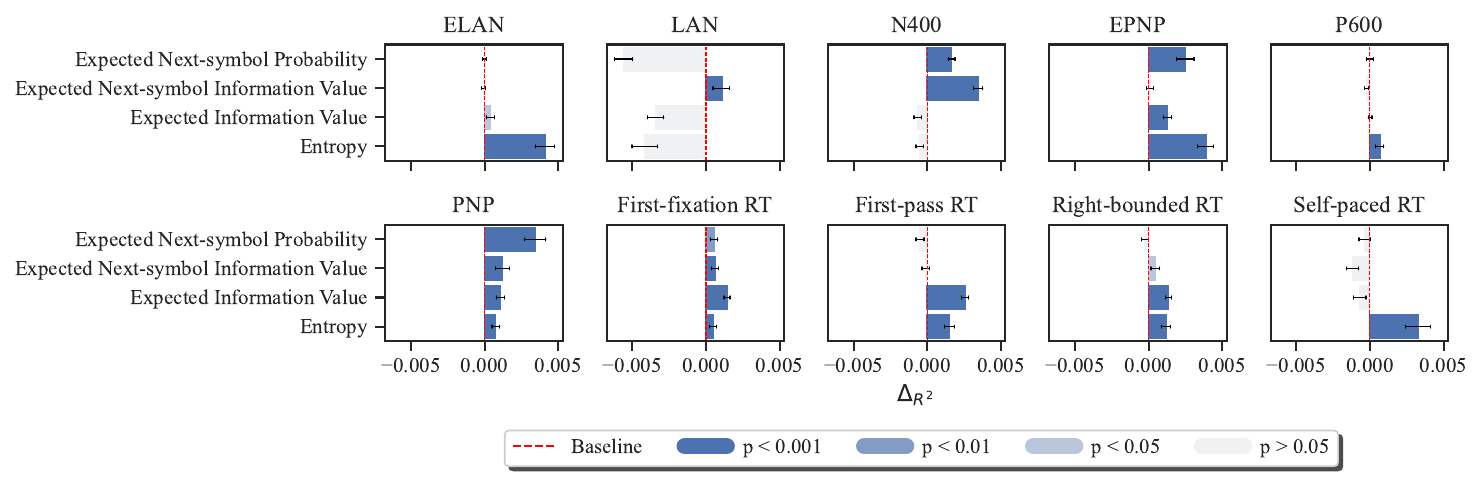}
		\caption{Difference between the predictive power of anticipatory measures used in combination with surprisal vs.\ expected next-symbol surprisal used in combination with surprisal; 95\% confidence intervals. The red dotted line represents the combined baseline regressor.
			Significance is color-coded, as described in the legend.
		}
		\label{fig:anticipatory-against-entropy}
		\vspace{-0.9em}
	\end{figure*}
	\subsection{Results}
	\label{sec:predictive-power-results}
	We now present our main results, obtained with \gpttwo{}, $\samplesize=2^9$, and $\maxseqlen=5$ tokens. 
	
	\subsubsection{Responsive Measures}
	\label{sec:results-responsive}
	The main results for responsive measures are visualized in \cref{fig:pred-power-responsive,fig:surprisal_prob_vs_ratings}. See \cref{fig:pred-power-responsive-gptneo,fig:probability-predictability-neo,fig:probability-predictability,fig:surprisal_prob_vs_ratings-gptneo} in \cref{sec:appendix-predictive-power-results} for further results.
	
	\paragraph{Cloze Completions and Predictability Ratings.}
	These are the two types of psycholinguistic data in the Aligned dataset that more explicitly quantify uncertainty for the upcoming unit.
	Indeed, predictive power here is 1-2 orders of magnitude larger than for ERPs and reading times (\cref{fig:pred-power-responsive} vs.\ \ref{fig:probability-predictability}).
	We find surprisal has the highest $\deltarsq$ for predictability ratings ($0.30 \pm 0.06$), while probability has stronger predictive power for human cloze completion probability ($0.48 \pm 0.07$).
	This result demonstrates the role of the warping function $\funcf$ in \Cref{eq:gen-surprisal} in fitting a given psycholinguistic construct or data type, even when the scoring function remains unchanged (see \cref{eq:surprisal-model-g,eq:probability-model-g}).
	As further illustrated in \Cref{fig:surprisal_prob_vs_ratings}, the same binary scoring function provides a good linear fit to human cloze probabilities but not to predictability ratings. 
	In contrast, logarithmic warping of the binary scores, which handles highly surprising outcomes more robustly, results in a better fit to predictability ratings.\looseness-1
	
	\paragraph{Event-related Brain Potentials.}
	Different responsive models---both in terms of warping and scoring function---align with different ERP components (\Cref{fig:pred-power-responsive}).
	Surprisal shows higher predictive power for EPNP and P600, probability is the best predictor of PNP amplitudes, and information value of LAN and N400.
	Information value's high predictive power for N400, a component believed to relate to semantic predictability \citep{brothers2020going}, may be explained in terms of this measure's semantically aware scoring function. 
	Furthermore, the predictive power of different responsive measures appears to correspond to groupings defined by the time windows in which the ERP amplitudes are recorded: LAN and N400 are detected roughly between 300 and 500 ms after the stimulus onset, EPNP and P600 occur between 400 and 700 ms, and PNP is the latest component, recorded between 600 and 700 ms after the onset~\cite{frank2015erp}.\footnote{
		In fact, a fourth cluster consists of ELAN, the earliest ERP component (125–175ms), for which only an anticipatory measure, entropy, shows significant predictive power (\cref{fig:anticipatory-against-entropy}).
	}
	Overall, this result highlights the importance of having a family of measures at disposal for targeted modeling of various psycholinguistic data.\looseness-1
	
	\paragraph{Reading Times.}
	For both eye-tracked and self-paced reading time data, surprisal demonstrates superior performance among responsive measures.
	As shown in \Cref{fig:pred-power-responsive}, probability is the second best predictor for eye-tracked reading times, and information value ranks second for self-paced reading times.
	These results imply that, among the competing theories we examined, surprisal theory---which posits that cognitive cost is linked to the magnitude of incremental updates in mental representation---provides a better explanation for the traditional notion of cost captured by reading times.
	
	\subsubsection{Anticipatory Measures}
	\label{sec:results-anticipatory}
	The main results for anticipatory measures are shown in \cref{fig:anticipatory-against-entropy}. See also \cref{fig:pred-power-anticipatory,fig:pred-power-anticipatory-neo,fig:pred-power-anticipatory-cloze-entropy,fig:pred-power-anticipatory-cloze-entropy-neo,fig:anticipatory-against-entropy-neo} in \Cref{sec:appendix-predictive-power-results}.\looseness-1
	
	\paragraph{Cloze Completions.}
	The most direct quantification of contextual uncertainty in the Aligned dataset is the entropy derived from human cloze completions.
	Indeed, all anticipatory measures show the strongest predictive power when used in isolation for this data type; see \Cref{fig:pred-power-anticipatory-cloze-entropy} in \Cref{sec:appendix-predictive-power-results} (default baseline).
	Expected next-symbol surprisal and probability have equivalent predictive power for close entropy, followed by expected next-symbol information value.
	The sequence-level anticipatory measures exhibit lower predictive capacity, with entropy obtaining the lowest $\deltarsq$.
	This is not surprising, considering that cloze completions are composed of individual words.\looseness-1
	
	\paragraph{Event-related Brain Potentials.}
	Echoing our findings with responsive measures, expected next-symbol information value is the strongest predictor for LAN, and even more so for N400, which emphasizes the connection between N400 and information value's semantically informed scoring function.
	Entropy also deserves a special mention: It stands out as the sole predictive measure for ELAN, where no other measure---responsive or anticipatory---is significantly predictive. It is also the best predictor of EPNP, with predictive power equivalent to surprisal.
	A common feature of ELAN and EPNP amplitudes is that they are detected by EEG sensors on frontal scalp regions.
	
	\paragraph{Reading Times.}
	As shown in \cref{fig:pred-power-anticipatory}, where we use the default baseline, reading times are generally less strongly predicted by anticipatory measures in isolation than by responsive ones.
	However, when considering the combined baseline with surprisal and expected next-symbol surprisal (\cref{fig:anticipatory-against-entropy}), we find replacing expected next-symbol surprisal with another anticipatory measure increases predictive power across all data types.
	The improvements likely stem from the higher complementarity of these measures with surprisal (see \cref{sec:analysis-correlations}).\looseness-1
	
	\subsection{Main Findings}
	Our experiments, conducted across a comprehensive range of psycholinguistic data, demonstrate that different generalized surprisal models---both responsive and anticipatory---provide complementary fit across different data types. 
	For behavioral responses collected in the cloze task~\cite{taylor1953cloze}, next-symbol probability accurately captures the distribution of human productions, while human predictability ratings are better explained by next-symbol surprisal. 
	For ERP components, the choice of measures has an impact on predictive power for amplitudes recorded at different onsets. Information value is a stronger predictor of early-onset components, while probability and surprisal are more predictive for late-onset ones. 
	Next-symbol information value, both in its responsive and anticipatory form, is consistently the best predictor for N400, an ERP compoment often predicted using surprisal \cite{frank-etal-2013-word, michaelov2024n400} but also known to be associated with semantic uncertainty \citep{brothers2020going,lindborg2023semantic}.
	On the other hand, sequence-level entropy, a measure whose computation involves long-horizon simulations, is predictive of ERP components in the frontal regions of the scalp, which are thought to be implicated in cognitive or executive control \cite{alexander1989frontal,kandel2000principles,fedorenko2013broad}.
	Finally, reading times, both self-paced and eye-tracked, are best predicted by responsive measures, with surprisal emerging as the overall best predictor. 
	However, when comparing models that include surprisal alongside an anticipatory predictor, replacing \citeposs{pimentel-etal-2023-anticipation}'s expected next-symbol surprisal with one of our alternative anticipatory measures yields significant increases in predictive power across all studied reading time variables.\looseness-1
	
	\section{Conclusion}
	We introduced a generalization over classic information-theoretic measures of predictive uncertainty in online language processing.
	Our generalized surprisal framework subsumes both responsive and anticipatory measures, including established special cases, but providing a vocabulary and the formal tools for experimenters to design new measures and explain psycholinguistic data of interest.

	\section*{Limitations}
	There are several special cases of generalized surprisal that we did not include in our experiments to maintain focus, ensure the interpretability of our results, and keep the scope appropriate for a conference paper. 
	In \cref{sec:other-special-cases}, we provide a few examples \cite{rabovsky2018modelling,li2023heuristic,opedal2024role,meister-etal-2024-similarity}.
	
	Other limitations of our study concern the psycholinguistic data under analysis. 
	We consider only English data and native English speakers, and thus, can only draw conclusions about incremental processing of English as L1.
	Multilingual datasets exist \cite[e.g.,][]{siegelman2022expanding,celer2022} and should be used in future work to test our findings for other languages as well as speakers of English with a different L1.
	Furthermore, the linguistic contexts in the analyzed dataset consist of a single sentence. More experimentation is needed to assess the predictive power of our different measures with more complex linguistic contexts such as whole paragraphs and texts, e.g., with the Natural Stories corpus \citep{futrell-etal-2018-natural}, or sequences of conversational turns, which are known to modulate predictive uncertainty in non-trivial ways \cite{giulianelli-fernandez-2021-analysing,giulianelli-etal-2021-information,tsipidi2024curves}.
	Generally speaking, contexts are representations of the current state of the world and can include extra-linguistic information \cite{ankener2018visual,giulianelli-2022-towards}. 
	Future work should also study responsive and anticipatory linguistic processing modulated by visual cues.
	For visuo-linguistic contexts, estimates of our generalized formula can be calculated using image-conditioned or video-conditioned LMs.
	
	Finally, while we experiment with increasing the sample size in \cref{sec:analysis-variance}, there could be other, more efficient ways to reduce variance. 
	Future work may tackle variance reduction through, for example, importance sampling from altered (e.g., temperature-annealed) language model distributions.
	
	\section*{Acknowledgments}
	Mario Giulianelli was supported by an ETH Zurich Postdoctoral Fellowship. Andreas Opedal received funding from the Max Planck ETH Center for Learning Systems. We thank the anonymous ARR reviewers for their insightful feedback, and Sarenne Wallbridge for early discussions on the relationship between standard surprisal and sampling-based measures.
	
	\bibliography{custom}
	
	\clearpage
	
	\appendix
	\onecolumn
	
	\section{Details of the Experimental Setup}
	\label{sec:experiments-details}
	
	\subsection{Measurements in the Aligned Dataset}
	\label{sec:aligned-data}
	In our experiments, we use the Aligned dataset \cite{devarda-etal-2023}, which consists of $\datasetsize = 1726$ target--context pairs from English novels annotated with several different neural and behavioral measurements. We present all types of measurements below.
	
	\paragraph{Cloze Completions.}
	In this incremental version of the cloze task \citep{taylor1953cloze}, participants are shown a sentence fragment $\ctx$ where the upcoming target word $\str$ is masked, and they are asked to guess what that target word will be. 
	Two data types are derived from cloze completions: the \textbf{cloze probability} of the upcoming word, estimated as the Laplace-smoothed (with pseudocount $\alpha = 1$) proportion of participants who pick that word, and the entropy of the Laplace-smoothed cloze distribution (\textbf{cloze entropy}).
	
	\paragraph{Predictability Ratings.} 
	Participants are presented with a sentence fragment $\ctx$ as well as its corresponding target $\str$.
	Then, they are asked to rate how likely they think the target is on a Likert scale from 1 to 5 \citep{delong2014predictability}. 
	Predictability ratings facilitate the analysis of low-probability words unlikely to appear among cloze completions.
	
	\paragraph{Event-related Brain Potentials (ERPs).}
	ERPs are small voltages generated by participants' neural activity and recorded via an electroencephalogram~\citep{donchin1979event}.
	Participants are tasked with reading a sentence, then their ERPs are post-processed to obtain word-level measurements, where each word in turn is the target $\str$ and its preceding words form the context $\ctx$.
	For further details, see \citet{frank2015erp}.
	The ERP components analyzed here are the \textbf{N400}, often associated with a word’s semantic predictability \citep{brothers2020going}, \textbf{P600}, implicated in syntactic integration processes \citep{Kaan2000ThePA}, (Early) Left Anterior Negativity (\textbf{ELAN} and \textbf{LAN}), linked to syntactic expectations and working memory \citep{Friederici2007MappingSF}, and (Early) Post-N400 Positivity (\textbf{EPNP} and \textbf{PNP}), thought to reflect lexical expectations \citep{thornhill2012lexical}. 
	
	\paragraph{Eye-tracked Reading Times.} 
	Participants read a sentence and the time spent looking at each word is recorded, so that each word, in turn, is the target $\str$ and its preceding words form the context $\ctx$.
	The Aligned dataset contains four types of eye-tracked reading time indices: \textbf{first-fixation time}, \textbf{first-pass time}, 
	\textbf{right-bounded time},  
	and \defn{go-past} time. 
	For more details, see \citet{frank2013reading}. 
	We exclude go-past time, a measurement that includes regressions to previous words, and was found to be noisy in this dataset \cite{devarda-etal-2023}.
	
	\paragraph{Self-paced Reading Times.}
	Participants read a sentence one word at a time in a stationary-window paradigm~\citep{just1982paradigms}.
	The time elapsed between the presentation of a word and the participant's key press to proceed to the next word is the \textbf{self-paced reading time}.
	Contexts $\ctx$ here are taken to be the words preceding the current word $\str$, although these are not physically present on the participants' screen.
	
	\subsection{Aggregating Multi-Token Estimates}
	\label{sec:appendix-multi-token}
	
	When a word is composed of multiple subword tokens, token-level estimates are aggregated. 
	For probability-based measures, we naturally multiply token-level estimates following the chain rule.
	For surprisal and information value, we sum token-level estimates (which, for surprisal, is equivalent to multiplying token probabilities).
	See \citet{giulianelli-etal-2024-proper} for a discussion on the proper treatment of tokenization in computational psycholinguistics.
	
	\section{Correlation Between Measures}
	\label{sec:analysis-correlations}
	To understand the potential overlap or complementarity between measures, we calculate their correlation. 
	\cref{sec:pearson-correlation} and \cref{sec:spearman-correlation} below present results in terms of Pearson correlation and Spearman rank-correlation, respectively.
	
	\subsection{Pearson Correlation} \label{sec:pearson-correlation}
\begin{figure*}[b!]
		\centering
		\includegraphics[width=0.95\linewidth]{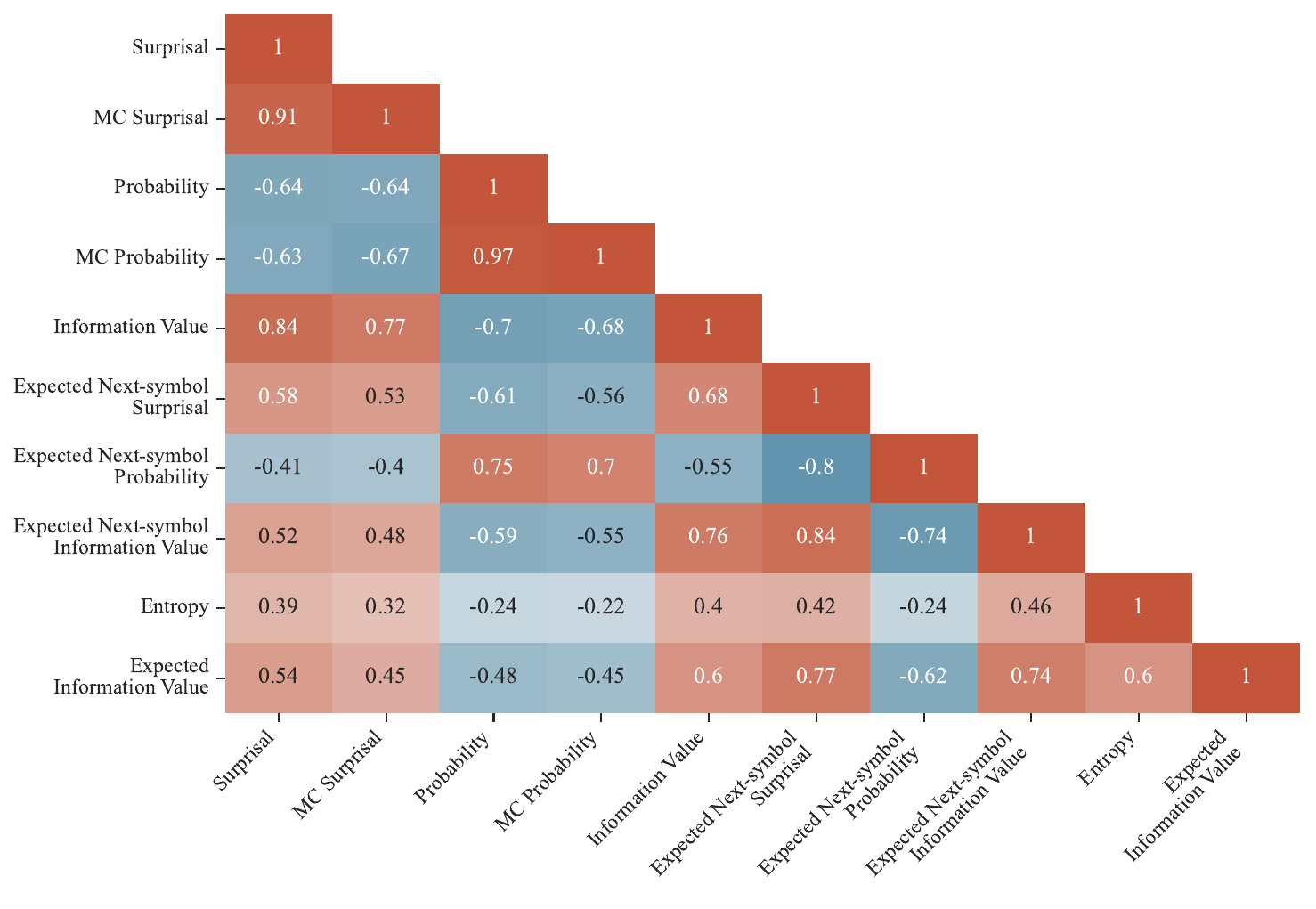}
		\caption{Pearson correlation between responsive and anticipatory measures. Estimates obtained for the Aligned dataset. Monte Carlo (MC) samples with $\samplesize=2^{9}$ and $\maxseqlen = 5$ from \gpttwo{}.}
		\label{fig:correlation_matrix_all_gpt2}
	\end{figure*}

\begin{figure*}
		\centering
		\includegraphics[width=0.95\linewidth]{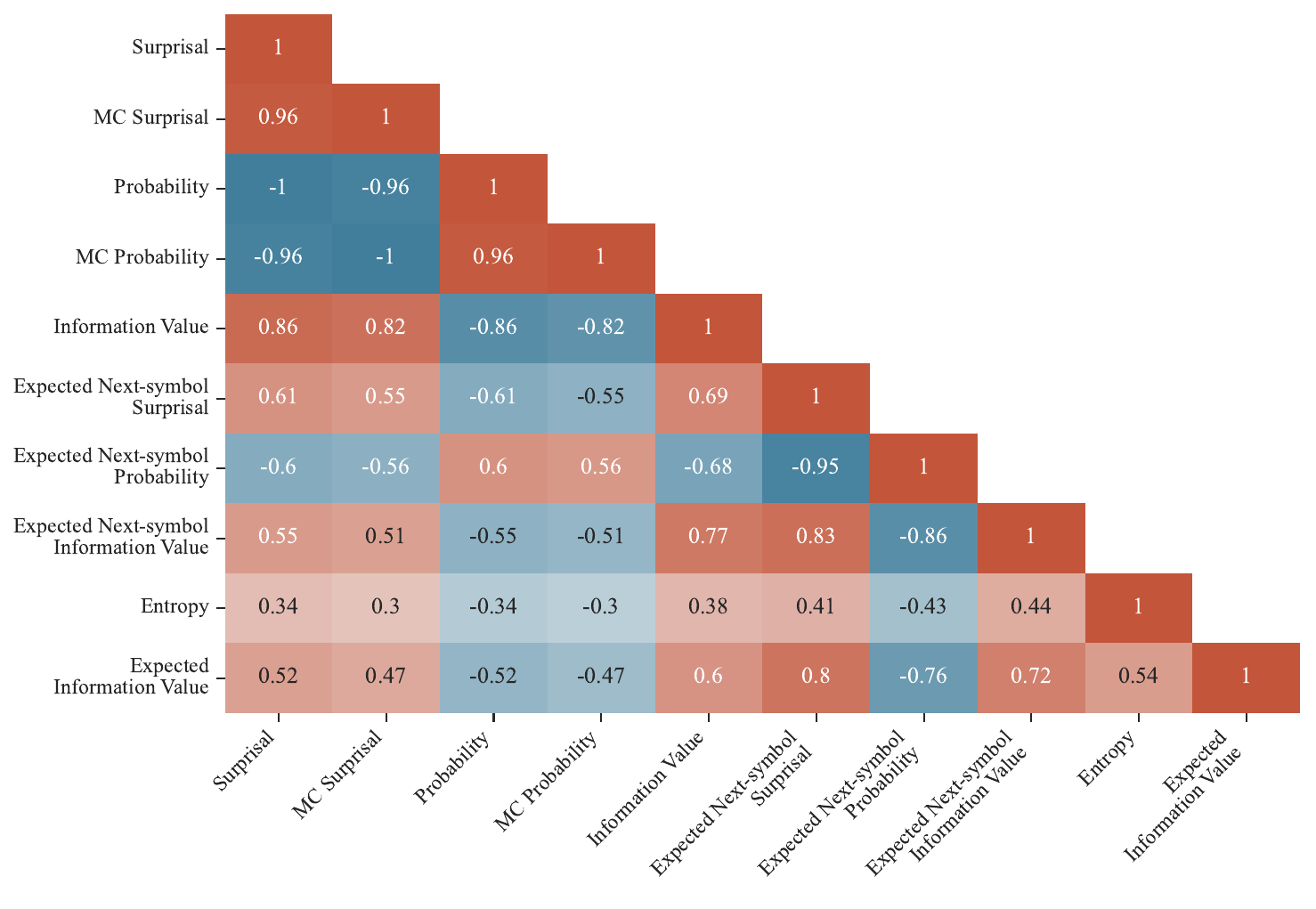}
		\caption{Spearman rank-correlation between responsive and anticipatory measures. Estimates obtained for the Aligned dataset. Monte Carlo (MC) samples with $\samplesize=2^{9}$ and $\maxseqlen = 5$ from \gpttwo{}.}
		\label{fig:spearman_correlation_matrix_all_gpt2}
	\end{figure*}

	\cref{fig:correlation_matrix_all_gpt2} shows Pearson correlation coefficients between estimates obtained with responsive and anticipatory measures.
	Estimates are computed on the 1726 prefix-continuation pairs of the Aligned dataset (\Cref{sec:exp-setup-data}) using \gpttwo{}, $\samplesize=2^{9}$ samples and a maximum sample length of $\maxseqlen = 5$ tokens.
	Results for \gptneo{}, which follow the same trends, are shown in \cref{fig:spearman_correlation_matrix_all_gpt2}.
	
\begin{figure*}
		\centering
		\includegraphics[width=0.95\linewidth]{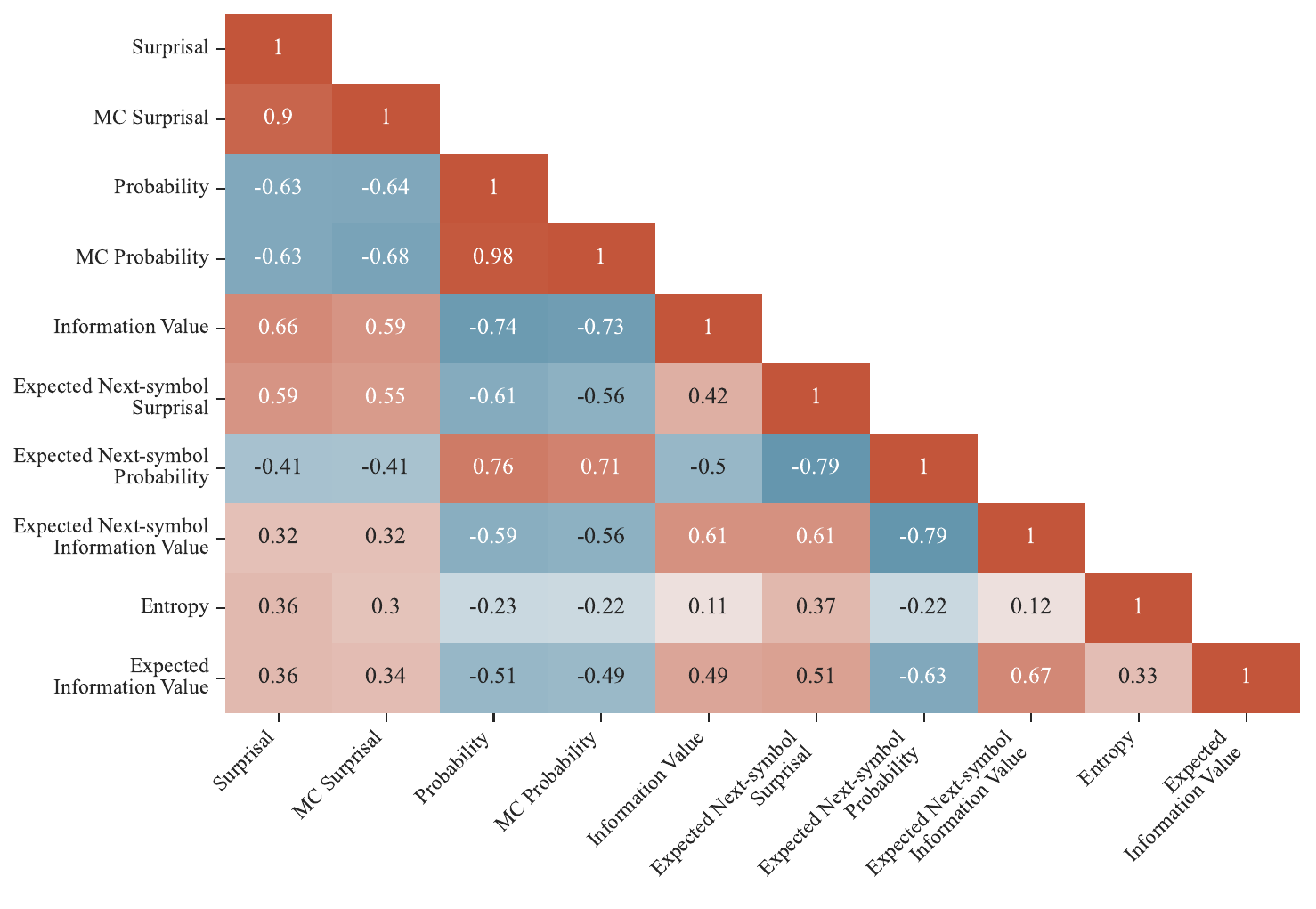}
		\caption{Pearson correlation between responsive and anticipatory measures. Estimates obtained for the Aligned dataset. Monte Carlo (MC) samples with $\samplesize=2^{9}$ and $\maxseqlen = 5$ from \gptneo{}.}
		\label{fig:correlation_matrix_all_gptneo}
	\end{figure*}

	First, we compare MC estimates of surprisal and probability to values computed in closed form, putting our theoretical argument in \Cref{sec:framework-generalizing} to the test. 
	Probability and MC probability have an almost perfect Pearson correlation ($r = 0.97$). 
	The correlation between surprisal and MC surprisal is also strong, although with a lower coefficient ($r = 0.91$).\footnote{
		We add a small constant ($1\mathrm{e}{-4}$) to the expected score in \Cref{eq:generalized-surprisal} before taking the logarithm to avoid numerical errors.
		The Pearson correlation coefficient is mildly sensitive to the choice of constant.
	} 
	The rank-correlation between the two pairs of measures is the same, $\rho = 0.96$, as shown in \Cref{fig:spearman_correlation_matrix_all_gpt2,fig:correlation_matrix_all_gptneo}.
	This result empirically confirms the formal argument put forward in \Cref{sec:framework-generalizing}: surprisal and probability can be expressed as expectations over continuations of partial linguistic stimuli, scored with an indicator function.
	
	Next, we investigate the relationship between measures of the same type, anticipatory or responsive.
	This allows us to evaluate whether different models of anticipatory and responsive processing lead to similar, diverging, or complementary measurements.
	We find that information value correlates more strongly with surprisal than with probability and, in line with this observation, that the highest correlation for anticipatory measures is between expected next-symbol surprisal and expected next-symbol information value.
	Expected next-symbol probability and expected next-symbol surprisal also correlate strongly ($r=0.80$).
	On the other hand, sequence-level anticipatory measures exhibit lower correlations overall; entropy, in particular, correlates only moderately with other measures.
	
	Finally, the anticipatory measure that correlates most strongly with surprisal is expected next-symbol surprisal (see both \cref{fig:correlation_matrix_all_gpt2,fig:spearman_correlation_matrix_all_gpt2} for \gpttwo{} and \cref{fig:correlation_matrix_all_gptneo,fig:spearman_correlation_matrix_all_gptneo} for \gptneo{}, where this is even more evident).
	This result contributes to explaining our findings for reading times in \cref{sec:results-anticipatory}, where we show that replacing expected next-symbol surprisal with another anticipatory measure, in a model that includes surprisal along with the default baselines, yields an increase in predictive power across all reading time variables.
	
	In summary, this correlation analysis provides empirical support to the theoretical foundations of our generalized framework and confirms that its different special cases quantify alternative notions of responsive and anticipatory processing.

	\subsection{Spearman Rank-Correlation}
	\label{sec:spearman-correlation}
\begin{figure*}
		\centering
		\includegraphics[width=0.95\linewidth]{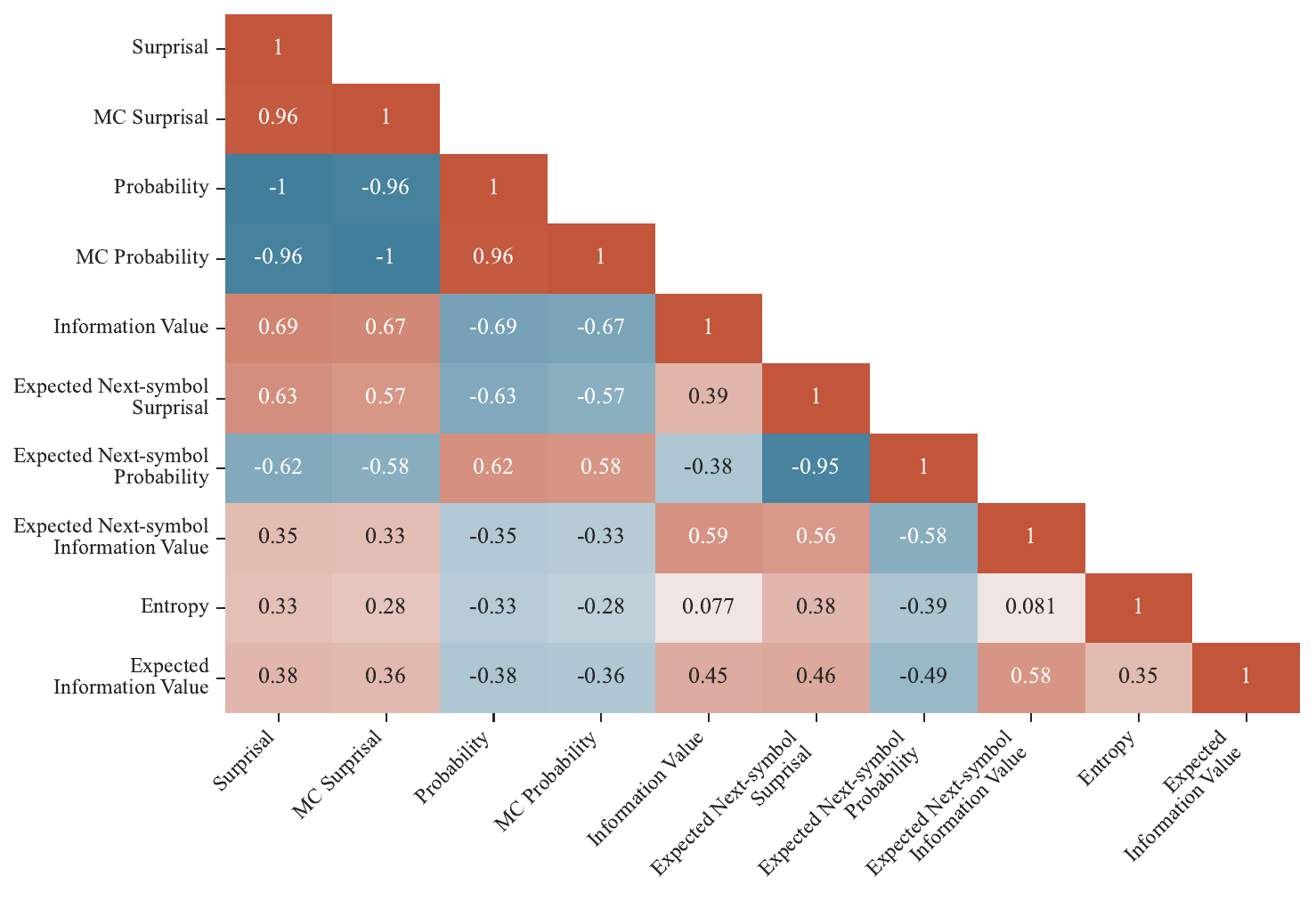}
		\caption{Spearman rank-correlation between responsive and anticipatory measures. Estimates obtained for the Aligned dataset. Monte Carlo (MC) samples with $\samplesize=2^{9}$ and $\maxseqlen = 5$ from \gptneo{}.}
		\label{fig:spearman_correlation_matrix_all_gptneo}
	\end{figure*}

	As a complement to \cref{sec:pearson-correlation}, we show the Spearman rank-correlation coefficients between responsive and anticipatory measures in \Cref{fig:spearman_correlation_matrix_all_gpt2,fig:spearman_correlation_matrix_all_gptneo}. Note the almost perfect correlation between surprisal and MC surprisal, identical to that of probability and MC probability.

	\section{Variation and Runtime Analysis}\label{sec:appendix-variance-runtime}
\begin{figure*}
		\centering
		\includegraphics[width=0.95\linewidth]{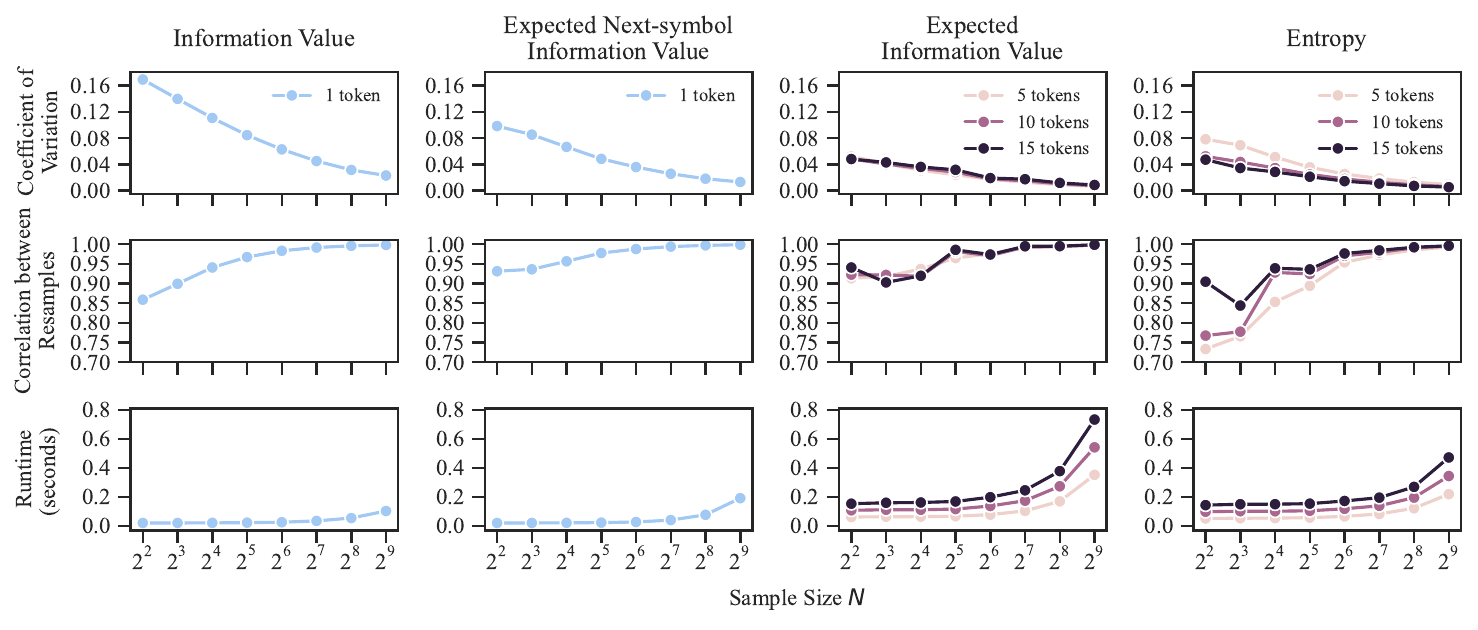}
		\vspace{-0.7em}
		\caption{\textit{Coefficient of variation} (top), \textit{correlation between resamples} (center), and \textit{runtimes} (bottom) for sampling-based measures across the stimuli in the Aligned dataset, using \gptneo{} as a language model.
			Confidence intervals ($95\%$) are too narrow to be visible; the horizontal axis is in log scale.}
		\label{fig:analysis-all-in-one-gptneo}
	\end{figure*}

	In \cref{sec:analysis} of the main paper, we presented an empirical analysis of Monte Carlo estimation, with the goal of understanding the trade-off between the variance and runtime of each measure's estimator. 
	Here, in \cref{fig:analysis-all-in-one-gptneo}, we display the result of this analysis conducted using \gptneo{}.
	
	\section{Psycholinguistic Predictive Power}
	\label{sec:appendix-predictive-power}
	
	\subsection{Statistical Analysis}
	\label{sec:appendix-significance-testing}
	To evaluate the predictive power of a generalized surprisal model, we use the following procedure.
	First, we run 10-fold cross-validation, iteratively partitioning the Aligned dataset into a 90\% training set and a 10\% test set and measuring the coefficients of determination $\rsq$ of the baseline and target regressors on the test set.
	We repeat this procedure using 100 random seeds, and collect the $\deltarsq$ scores associated with the target predictor.
	These are the scores that determine the width of the bars and the confidence intervals in \cref{fig:probability-predictability,fig:probability-predictability-neo,fig:pred-power-responsive,fig:pred-power-responsive-gptneo,fig:pred-power-anticipatory,fig:pred-power-anticipatory-neo,fig:pred-power-anticipatory-cloze-entropy,fig:pred-power-anticipatory-cloze-entropy-neo,fig:anticipatory-against-entropy,fig:anticipatory-against-entropy-neo,fig:pred-power-anticipatory,fig:pred-power-anticipatory-neo}.
	Then, to assess the significance of a measure's predictive power---i.e., of a measure's positive $\deltarsq$ scores---we run paired permutation tests\footnote{
		We use the implementation provided by the \texttt{SciPy} library under \texttt{\href{https://docs.scipy.org/doc/scipy/reference/generated/scipy.stats.permutation_test.html}{scipy.stats.permutation\_test}}.
	} under the null hypothesis that the target regressor's $\rsq$ is smaller or equal to the baseline regressor's $\rsq$ and the alternative hypothesis that the target regressor's $\rsq$ is greater than the baseline regressor's $\rsq$.
	We use 10,000 resamples and the difference between the sample means as a test statistic.
	The $p$-value output by the permutation test (as color-coded or indicated in the captions of \cref{fig:probability-predictability,fig:probability-predictability-neo,fig:pred-power-responsive,fig:pred-power-responsive-gptneo,fig:pred-power-anticipatory,fig:pred-power-anticipatory-neo,fig:pred-power-anticipatory-cloze-entropy,fig:pred-power-anticipatory-cloze-entropy-neo,fig:anticipatory-against-entropy,fig:anticipatory-against-entropy-neo,fig:pred-power-anticipatory,fig:pred-power-anticipatory-neo}) is the proportion of the randomized null distribution that is as extreme as the observed value of the test statistic.
	
	For comparison between pairs of regressors which both include target predictors, we use the same procedure, only considering the baseline regressor with each of the target regressors in turn.
	Our full analysis is implemented in Python and available at \url{https://github.com/rycolab/generalized-surprisal}.
	
	\subsection{Further Results}
	\label{sec:appendix-predictive-power-results}
\begin{figure*}[t!]
		\centering
		\includegraphics[width=0.99\linewidth]{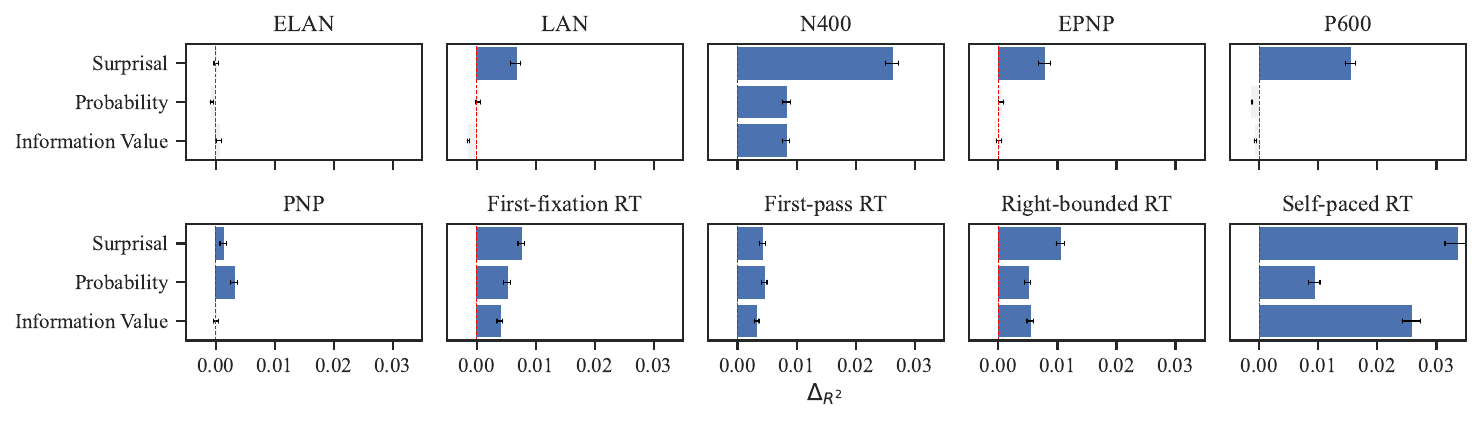}
		\vspace{-1.0em}
		\caption{Predictive power $\deltarsq$ of responsive generalized surprisal models for event-related potentials and reading times, using \gptneo{} as a language model; 95\% confidence intervals. Significance color-coded: blue for $p<0.0001$, gray for $p>0.01$.\looseness-1}
		\label{fig:pred-power-responsive-gptneo}
	\end{figure*}

\begin{figure}[t]
		\centering
		\includegraphics[width=0.55\linewidth]{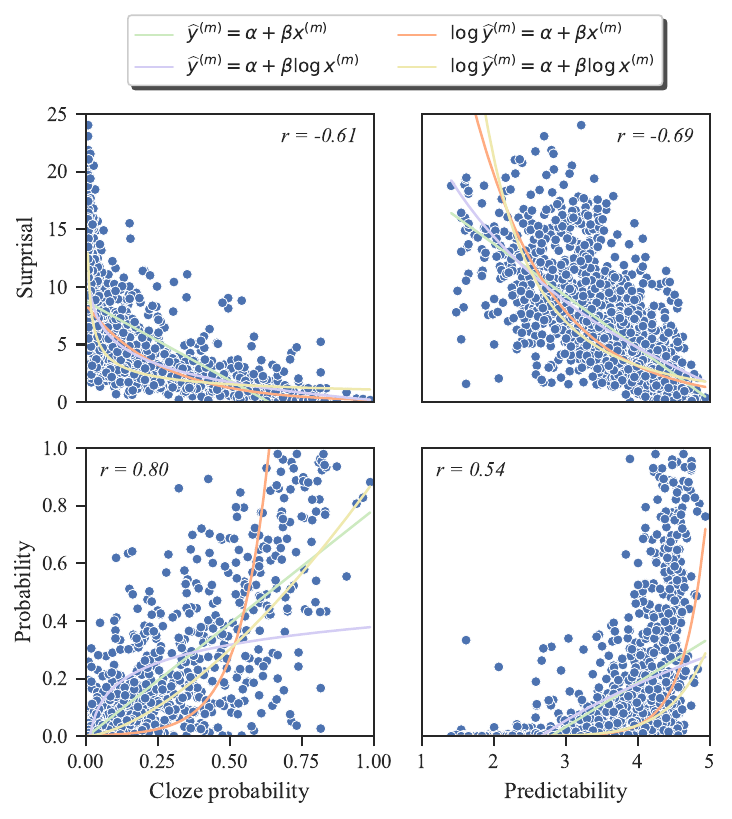}
		\vspace{-0.5em}
		\caption{Probability and surprisal (\gptneo{}) against human cloze probabilities and predictability ratings, with Pearson correlation coefficients $r$. For regressions, $x^{(\datasetidx)} =  \wordmeasure(\str^{(\datasetidx)}, \ctx^{(\datasetidx)})$ is the predictor and $y^{(\datasetidx)} = \datum(\str^{(\datasetidx)}, \ctx^{(\datasetidx)})$ the predicted variable.
		}
		\label{fig:surprisal_prob_vs_ratings-gptneo}
	\end{figure}

\begin{figure*}
		\centering
		\includegraphics[width=0.65\linewidth]{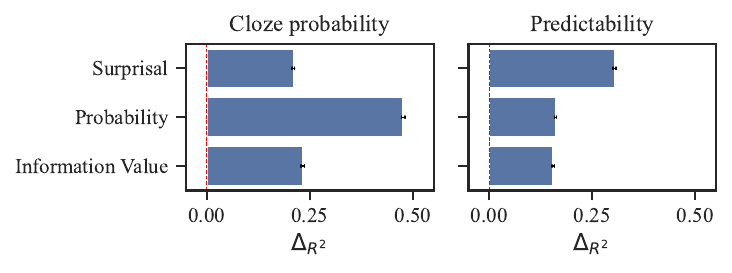}
		\vspace{-0.5em}
		\caption{Predictive power $\deltarsq$ of responsive measures for human cloze probabilities and predictability ratings, using \gpttwo{} as a language model; 95\% confidence intervals. The red dotted line represents the default baseline regressor. All measures are significantly predictive ($p<0.001$).}
		\label{fig:probability-predictability}
	\end{figure*}

\begin{figure*}
		\centering
		\includegraphics[width=0.65\linewidth]{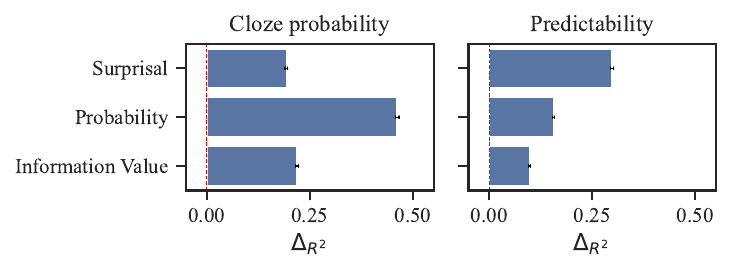}
		\vspace{-0.5em}
		\caption{Predictive power $\deltarsq$ of responsive measures for human cloze probabilities and predictability ratings, using \gptneo{} as a language model; 95\% confidence intervals. The red dotted line represents the default baseline regressor. All measures are significantly predictive ($p<0.001$).}
		\label{fig:probability-predictability-neo}
	\end{figure*}

	\paragraph{Responsive Measures.}
	To complement \cref{fig:pred-power-responsive,fig:surprisal_prob_vs_ratings} in the main paper, \cref{fig:pred-power-responsive-gptneo,fig:surprisal_prob_vs_ratings-gptneo} show the results for responsive measures obtained with \gptneo{}.
	\Cref{fig:probability-predictability,fig:probability-predictability-neo} show the $\deltarsq$ scores of our three responsive measures for cloze probability and predictability.
	
\begin{figure*}
		\centering
		\includegraphics[width=0.8\linewidth]{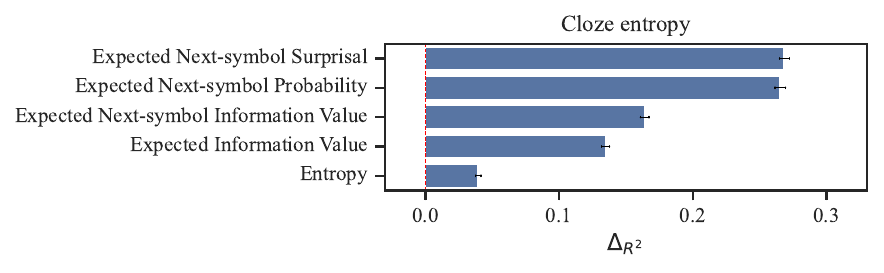}
		\caption{Psycholinguistic predictive power $\deltarsq$ of anticipatory generalized surprisal models for the cloze entropy of the human cloze completion distributions, using \gpttwo{} as a language model; 95\% confidence intervals. The red dotted line represents the default baseline regressor. All measures have significant predictive power ($p<0.0001$).}
		\label{fig:pred-power-anticipatory-cloze-entropy}
	\end{figure*}

\begin{figure*}
		\centering
		\includegraphics[width=0.8\linewidth]{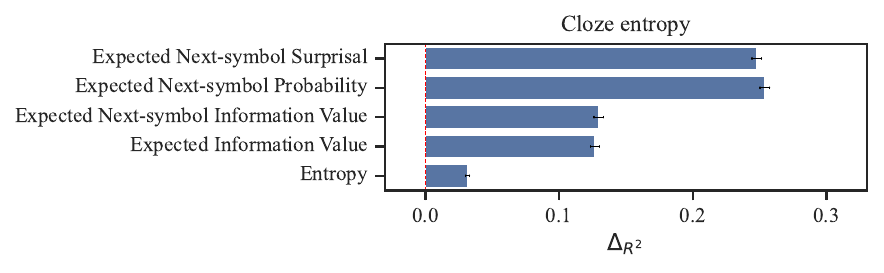}
		\caption{Psycholinguistic predictive power $\deltarsq$ of anticipatory generalized surprisal models for the cloze entropy of the human cloze completion distributions, using \gptneo{} as a language model; 95\% confidence intervals. The red dotted line represents the default baseline regressor. All measures have significant predictive power ($p<0.0001$).}
		\label{fig:pred-power-anticipatory-cloze-entropy-neo}
	\end{figure*}

\begin{figure*}
		\centering
		\includegraphics[width=0.98\linewidth]{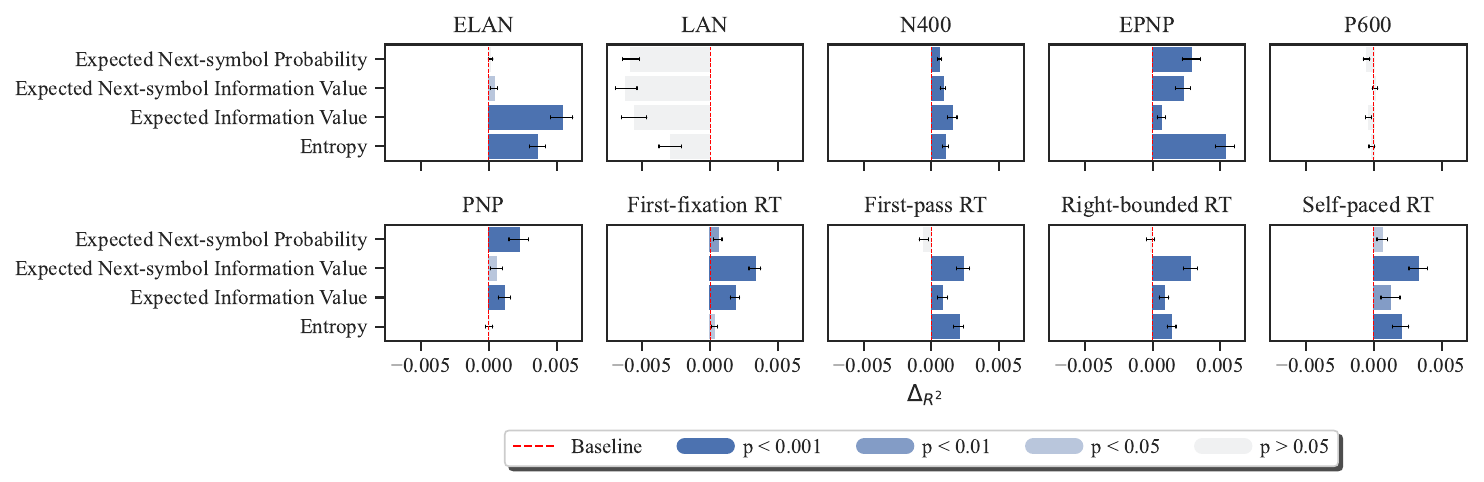}
		\caption{Difference between the predictive power of anticipatory measures used in combination with surprisal vs.\ expected next-symbol surprisal in combination with surprisal; 95\% confidence intervals. \gptneo{} is used as the language model. The red dotted line represents the combined baseline regressor.
			Significance is color-coded, as described in the legend.
		}
		\label{fig:anticipatory-against-entropy-neo}
	\end{figure*}

	\paragraph{Anticipatory Measures.}
	To complement \cref{fig:anticipatory-against-entropy} in the main paper, \cref{fig:anticipatory-against-entropy-neo} shows the results for anticipatory measures used in combination with surprisal, with \gptneo{} estimates.
	\Cref{fig:pred-power-anticipatory-cloze-entropy,fig:pred-power-anticipatory-cloze-entropy-neo} show the predictive power of anticipatory measures for cloze entropy.
	
\begin{figure*}[h!]
		\centering
		\includegraphics[width=\linewidth]{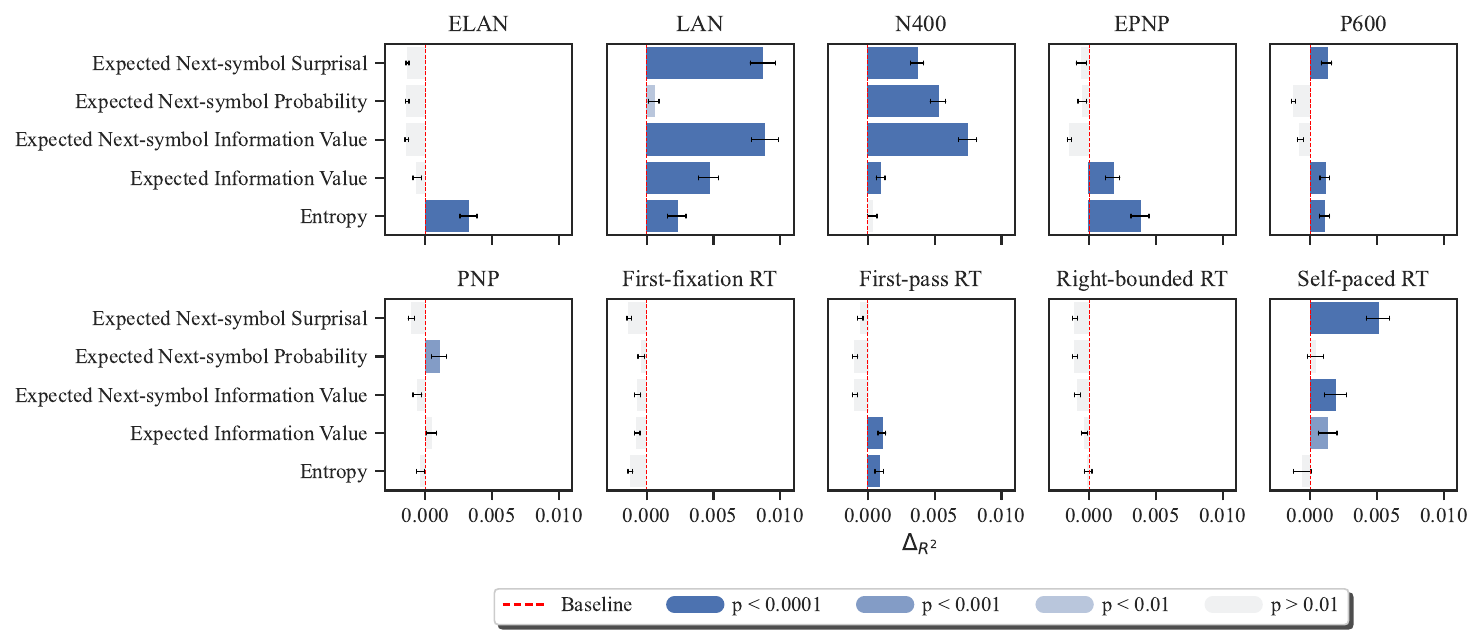}
		\caption{Psycholinguistic predictive power $\deltarsq$ of anticipatory generalized surprisal models for event-related potentials and reading times, using \gpttwo{} as a language model; 95\% confidence intervals. The red dotted line represents the default baseline regressor. Significance is color-coded, as described in the legend.}
		\label{fig:pred-power-anticipatory}
	\end{figure*}

\begin{figure*}[h!]
		\centering
		\includegraphics[width=\linewidth]{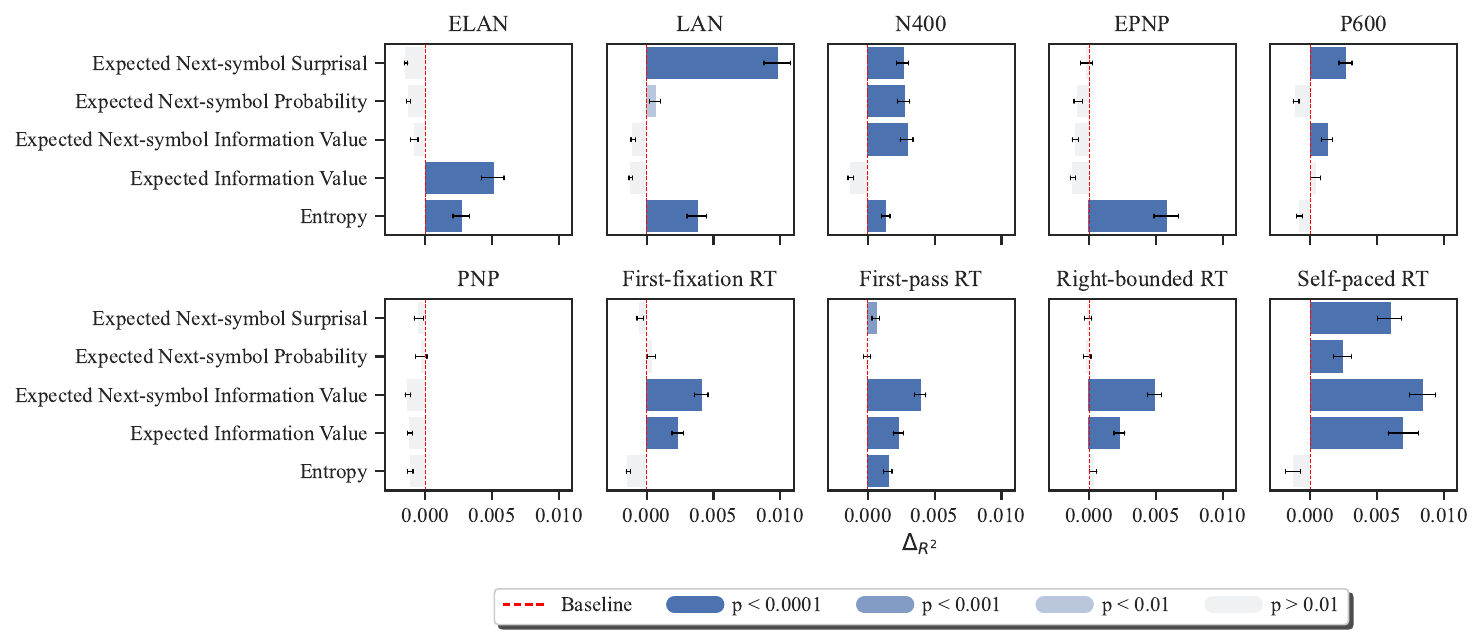}
		\caption{Psycholinguistic predictive power $\deltarsq$ of anticipatory generalized surprisal models for event-related potentials and reading times, using \gptneo{} as a language model; 95\% confidence intervals. The red dotted line represents the default baseline regressor. Significance is color-coded, as described in the legend.}
		\label{fig:pred-power-anticipatory-neo}
	\end{figure*}

	In \cref{sec:results-anticipatory} of the main paper (\cref{fig:anticipatory-against-entropy}), we evaluate anticipatory measures against a combined baseline that includes surprisal and expected next-symbol surprisal next to the default baseline variables (target length, target frequency, and prefix length).
	Here, in \cref{fig:pred-power-anticipatory,fig:pred-power-anticipatory-neo}, we show the predictive power results for the anticipatory measures against the default baseline.

	\section{Other Special Cases of Generalized Surprisal} \label{sec:other-special-cases}
	As noted in the Limitations section, there are several special cases of generalized surprisal that we did not include in our experiments to maintain a focused scope for the paper. Below, we provide a few examples.
	
	\paragraph{Semantic Update.} This model, proposed by \citet{rabovsky2018modelling}, is based on changes in neural representations and is given by:
	\boxalign{
		\begin{subequations}
			\begin{align}
				\funcf(x) = &\ x \\
				\funcg(\continuation, \str, \ctx) =& 
				\indicator{\sympos{1} \preceq  \continuation} \sum_{i \in \mathcal I} |a_i(\sympos{1}) - a_i(\symc_{|\ctx|})|, 
			\end{align}
		\end{subequations}
	}
	in which $\mathcal I$ is an index set corresponding to the neurons at some particular layer in a neural network implementation of a language model, and $a_i(\symu)$ represents the sigmoid activation of neuron $i$ for symbol~$\symu$.

	\paragraph{Pointwise Mutual Information.} The pointwise mutual information between a word and its context, which under certain conditions yields expressive power equivalent to surprisal \citep{opedal2024role}, can be written as:
	\boxalign{
		\begin{subequations}
			\begin{align}
				\funcf(x) &= \log(x) \\
				\funcg(\continuation, \str, \ctx) &= \pLM(\ctx) \cdot \indicator{\sympos{1} \preceq \continuation}. 
			\end{align}
		\end{subequations}
	}
	
	\paragraph{Similarity-adjusted Surprisal.} The similarity-adjusted notion of surprisal proposed by \citet{meister-etal-2024-similarity} is analogous to information value but uses a similarity function $\similarity_{\ctx} \colon \kleene{\alphabet} \times \kleene{\alphabet} \to [0,1]$ as a scoring function and the negative logarithm as a warping function, as given by the following model: 
	\boxalign{
		\begin{subequations}
			\begin{align}
				\funcf(x) &= - \log(x) \\
				\funcg(\continuation, \str, \ctx) &= \similarity_{\ctx}(\continuation, \str)
			\end{align}
		\end{subequations}
	}
	
	Lastly, we note that the decomposition introduced by \citet{Li2023decomposition} is mathematically equivalent to surprisal and is therefore also captured by our framework.\looseness=-1
	
\end{document}

%% file: macros.tex
\usepackage[disable]{todonotes}
\newcommand{\note}[4][]{\todo[author=#2,color=#3,size=\scriptsize,fancyline,caption={},#1]{#4}} 

\newcommand{\ryan}[2][]{\note[#1]{\textbf{Ryan}}{green!30}{#2}}

\newcommand{\response}[2]{\vspace{3pt}\hrule\vspace{3pt}\textbf{#1: }{#2}}

\usepackage{colortbl}
\definecolor{ETHBlue}{RGB}{33,92,175}   
\definecolor{ETHGreen}{RGB}{98,115,19}      
\definecolor{ETHPurpleDark}{RGB}{140,10,89} 
\definecolor{ETHPurple}{RGB}{163,7,116} 
\definecolor{ETHGray}{RGB}{111,111,111} 
\definecolor{ETHRed}{RGB}{183,53,45}    
\definecolor{ETHPetrol}{RGB}{0,120,148} 
\definecolor{ETHBronze}{RGB}{142,103,19}    
\definecolor{ETHOrange}{RGB}{230, 100, 50}

\colorlet{MacroColor}{ETHPetrol}
\colorlet{MACROCOLOR}{MacroColor}

\usepackage[capitalize,noabbrev]{cleveref}
\crefname{section}{\S}{\S\S}
\crefname{table}{Tab.}{Tab.}
\crefname{figure}{Fig.}{Figs.}
\crefname{algorithm}{Alg.}{}
\crefname{equation}{Eq.}{Eq.}
\crefname{appendix}{App.}{}
\crefname{theorem}{Theorem}{}
\crefname{prop}{Proposition}{}
\crefname{definition}{Def.}{}
\crefname{cor}{Corollary}{}
\crefname{observation}{Observation}{}
\crefname{assumption}{Assumption}{}
\crefname{hyp}{Hyp.}{Hypotheses}
\crefformat{section}{\S#2#1#3}
\crefname{namedtheorem}{Hyp.}{Hypotheses}

\usepackage{xcolor}
\newcommand{\mvar}{}

\newtheorem{definition}{Definition}

\newcommand{\gpttwo}{GPT-2 Small} 
\newcommand{\gptneo}{GPT-Neo 125M}
\newcommand{\repr}{{\mvar \mathrm{r}}}
\newcommand{\embeddim}{{\mvar D}}

\newcommand{\rsq}{{\mvar R^2}}
\newcommand{\deltarsq}{{\mvar \Delta_{\rsq}}}

\newcommand{\samplesize}{{\mvar N}}
\newcommand{\sampleidx}{{\mvar n}}
\newcommand{\datasetsize}{{\mvar M}}
\newcommand{\datasetidx}{{\mvar m}}
\newcommand{\bootstrapsamples}{{\mvar B}}
\newcommand{\maxseqlen}{{\mvar L}}

\newcommand{\str}{{\mvar \boldsymbol{w}}}
\newcommand{\ctx}{{\mvar\boldsymbol{c}}}
\newcommand{\strpr}{{\mvar\boldsymbol{w'}}}

\newcommand{\kleene}[1]{{\mvar #1^*}}

\newcommand{\pLM}{{\mvar p}}
\newcommand{\paLM}{{\mvar \pi_{\pLM}}}
\newcommand{\qLM}{{\mvar p}}
\newcommand{\qaLM}{{\mvar \pi_{\qLM}}}

\newcommand{\sym}{{\mvar w}}
\newcommand{\symc}{{\mvar c}}

\newcommand{\sympos}[1]{{\mvar \sym_{#1}}}

\newcommand{\alphabet}{{\mvar \Sigma}}

\newcommand{\surprisal}{{\mvar \iota}}

\newcommand{\wordmeasure}{{\mvar \gamma}}
\newcommand{\estwordmeasure}{{\mvar \widehat{\wordmeasure}}}

\newcommand{\datum}{{\mvar \psi}}
\newcommand{\funcf}{{\mvar \mathrm{f}}}
\newcommand{\funcg}{{\mvar \mathrm{g}}}
\newcommand{\defeq}[0]{\,{\mvar \mathrel{\stackrel{\textnormal{\tiny def}}{=}}\,}}

\newcommand{\continuation}{{\mvar \boldsymbol{v}}}
\newcommand{\contsym}{{\mvar v}}
\newcommand{\contpos}[1]{{\mvar \contsym_{#1}}}
\newcommand{\contposprime}[1]{{\mvar \contsym'_{#1}}}
\newcommand{\distance}{{\mvar \mathrm{d}}}
\newcommand{\similarity}{{\mvar \mathrm{z}}}
\DeclareMathOperator*{\Expect}{{\mvar \mathbb{E}}}

\newcommand{\indicator}[1]{{\mvar \mathbbm{1}\{#1\}}}
\newcommand{\symu}{{\mvar u}}

\newcommand{\defn}[1]{\textbf{#1}}

\definecolor{darkblack}{rgb}{0.0,0.0,0.5}
\definecolor{darkgreen}{rgb}{0.0, 0.42, 0.24}
\definecolor{lightgreen}{rgb}{0.52, 0.73, 0.4}
\definecolor{darkgray}{rgb}{0.4,0.4,0.4}
\definecolor{darkblue}{rgb}{0.0,0.0,0.5}
\definecolor{darkpurple}{rgb}{0.5,0.2,0.8}
\definecolor{lightpurple}{rgb}{0.8,0.5,1}
\definecolor{verydarkgray}{rgb}{0.35,0.35,0.35}

\usepackage{tikz}
\newcommand{\boxalign}[2][0.97\linewidth]{
 \par\noindent\tikzstyle{mybox} = [fill=gray!10, inner sep=6pt]
 \begin{center}\begin{tikzpicture}
  \node [mybox] (box){%
   \begin{minipage}{#1}{\vspace{-5mm}#2}\end{minipage}
  };
 \end{tikzpicture}\end{center}
}

\usepackage[breakable, theorems, skins]{tcolorbox}
\tcbset{enhanced}

\newcommand{\pHum}{\mvar p_{\mathrm{H}}}